\documentclass{article}

\usepackage{arxiv}

\usepackage[utf8]{inputenc} 
\usepackage[T1]{fontenc}    
\usepackage{hyperref}       
\usepackage{url}            
\usepackage{booktabs}       
\usepackage{amsfonts}       
\usepackage{nicefrac}       
\usepackage{microtype}      
\usepackage{lipsum}		
\usepackage{graphicx}
\usepackage{natbib}
\usepackage{doi}
\usepackage{multirow}
\usepackage{subfigure}
\usepackage{amssymb}
\usepackage{amsmath}
\usepackage{setspace}
\usepackage{float}

\pdfoutput=1
\hypersetup{
    colorlinks=true,
    linkcolor=black,
    citecolor=black,
    filecolor=black,
    urlcolor=black,
}

\title{A CNN Approach to Simultaneously Count Plants and Detect Plantation-Rows from UAV Imagery}

\date{February, 2021 \\ arXiv:2012.15827 | Computer Science \\ Computer Vision and Pattern Recognition}	

\author{ \href{https://orcid.org/0000-0002-0258-536X}{\includegraphics[scale=0.06]{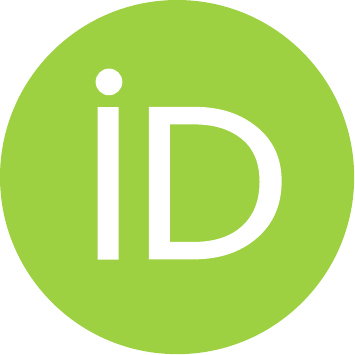}\hspace{1mm}Lucas Prado Osco}\thanks{Available at: \url{https://arxiv.org/abs/2012.15827}}\\
	University of Western São Paulo (UNOESTE)\\
	Presidente Prudente, SP, Brazil \\
	\texttt{lucasosco@unoeste.br} \\
	\And
	\href{https://orcid.org/0000-0002-2273-8968}{\includegraphics[scale=0.06]{orcid.pdf}\hspace{1mm}Mauro dos Santos de Arruda} \\
	Federal University of Mato Grosso do Sul (UFMS)\\
	Campo Grande, MS, Brazil \\
	\texttt{mauro.arruda@ufms.br} \\
	\And
	\href{https://orcid.org/0000-0002-4527-5724}{\includegraphics[scale=0.06]{orcid.pdf}\hspace{1mm}Diogo Nunes Gon\c{c}alves}\\
	Federal University of Mato Grosso do Sul (UFMS)\\
	Campo Grande, MS, Brazil \\
	\texttt{dnunesgoncalves@gmail.com} \\
	\And
	\href{https://orcid.org/0000-0000-0000-0000}{\includegraphics[scale=0.06]{orcid.pdf}\hspace{1mm}Alexandre Dias}\\
	Federal University of Mato Grosso do Sul (UFMS)\\
	Campo Grande, MS, Brazil \\
	\texttt{alexandre.dias@ufms.br} \\
	\And
	\href{https://orcid.org/0000-0000-0000-0000}{\includegraphics[scale=0.06]{orcid.pdf}\hspace{1mm}Juliana Batistot}\\
	Federal University of Mato Grosso do Sul (UFMS)\\
	Campo Grande, MS, Brazil \\
	\texttt{juliana.batistoti@ufms.br} \\
	\And
	\href{https://orcid.org/0000-0000-0000-0000}{\includegraphics[scale=0.06]{orcid.pdf}\hspace{1mm}Maurício de Souza}\\
	Federal University of Mato Grosso do Sul (UFMS)\\
	Campo Grande, MS, Brazil \\
	\texttt{mauricio.souza@ufms.br} \\
	\And
	\href{https://orcid.org/0000-0000-0000-0000}{\includegraphics[scale=0.06]{orcid.pdf}\hspace{1mm}Felipe David Georges Gomes} \\
	University of Western São Paulo (UNOESTE)\\
	Presidente Prudente, SP, Brazil \\
	\texttt{felipedgg@yahoo.com.br} \\
	\And	
	\href{https://orcid.org/0000-0001-6633-2903}{\includegraphics[scale=0.06]{orcid.pdf}\hspace{1mm}Ana Paula Marques Ramos} \\
	University of Western São Paulo (UNOESTE)\\
	Presidente Prudente, SP, Brazil \\
	\texttt{anaramos@unoeste.br} \\
	\And
	\href{https://orcid.org/0000-0001-8341-3203}{\includegraphics[scale=0.06]{orcid.pdf}\hspace{1mm}Lúcio André de Castro Jorge} \\
	Brazilian Agricultural Research Agency (EMBRAPA)\\
	São Carlos, SP, Brazil \\
	\texttt{lucio.jorge@embrapa.br} \\
	\And
	\href{https://orcid.org/0000-0003-0564-7818}{\includegraphics[scale=0.06]{orcid.pdf}\hspace{1mm}Veraldo Liesenberg} \\
	University of Santa Catarina State (UDESC)\\
	Santa Caratina, SC, Brazil \\
	\texttt{veraldo.liesenberg@udesc.br} \\
	\And
	\href{https://orcid.org/0000-0001-7899-0049}{\includegraphics[scale=0.06]{orcid.pdf}\hspace{1mm}Jonathan Li} \\
	University of Waterloo (UW)\\
	Waterloo, ON, Canada \\
	\texttt{junli@uwaterloo.ca} \\
	\And
	\href{https://orcid.org/0000-0000-0000-0000}{\includegraphics[scale=0.06]{orcid.pdf}\hspace{1mm}Lingfei Ma } \\
	University of Waterloo (UW)\\
	Waterloo, ON, Canada \\
	\texttt{l53ma@uwaterloo.ca} \\
	\And
	\href{https://orcid.org/0000-0002-9096-6866}{\includegraphics[scale=0.06]{orcid.pdf}\hspace{1mm}José Marcato Junior} \\
	Federal University of Mato Grosso do Sul (UFMS)\\
	Campo Grande, MS, Brazil \\
	\texttt{jose.marcato@ufms.br} \\
	\And
	\href{https://orcid.org/0000-0002-8815-6653}{\includegraphics[scale=0.06]{orcid.pdf}\hspace{1mm}Wesley Nunes Gonçalves} \\
	Federal University of Mato Grosso do Sul (UFMS)\\
	Campo Grande, MS, Brazil \\
	\texttt{wesley.goncalves@ufms.br} \\
}

\renewcommand{\shorttitle}{\textit{arXiv:2012.15827 - Osco et al. (2021)}}

\hypersetup{
pdftitle={A CNN Approach to Simultaneously Count Plants and Detect Plantation-Rows from UAV Imagery},
pdfsubject={cs.CV},
pdfauthor={Osco, et al.},
pdfkeywords={Deep learning, UAV imagery, Object detection, Remote sensing, Precision agriculture},
}

\begin{document}
\maketitle

\begin{abstract}
\textbf{Short (arXiv Abstract):} In this paper, we propose a novel deep learning method based on a Convolutional Neural Network (CNN) that simultaneously detects and geolocates plantation-rows while counting its plants considering highly-dense plantation configurations. The experimental setup was evaluated in a cornfield with different growth stages and in a Citrus orchard. Both datasets characterize different plant density scenarios, locations, types of crops, sensors, and dates. A two-branch architecture was implemented in our CNN method, where the information obtained within the plantation-row is updated into the plant detection branch and retro-feed to the row branch; which are then refined by a Multi-Stage Refinement method. In the corn plantation datasets (with both growth phases, young and mature), our approach returned a mean absolute error (MAE) of 6.224 plants per image patch, a mean relative error (MRE) of 0.1038, precision and recall values of 0.856, and 0.905, respectively, and an F-measure equal to 0.876. These results were superior to the results from other deep networks (HRNet, Faster R-CNN, and RetinaNet) evaluated with the same task and dataset. For the plantation-row detection, our approach returned precision, recall, and F-measure scores of 0.913, 0.941, and 0.925, respectively. To test the robustness of our model with a different type of agriculture, we performed the same task in the citrus orchard dataset. It returned an MAE equal to 1.409 citrus-trees per patch, MRE of 0.0615, precision of 0.922, recall of 0.911, and F-measure of 0.965. For citrus plantation-row detection, our approach resulted in precision, recall, and F-measure scores equal to 0.965, 0.970, and 0.964, respectively. The proposed method achieved state-of-the-art performance for counting and geolocating plants and plant-rows in UAV images from different types of crops. \\

\textbf{Complete (ISPRS Abstract):} Accurately mapping croplands is an important prerequisite for precision farming since it assists in field management, yield-prediction, and environmental management. Crops are sensitive to planting patterns and some have a limited capacity to compensate for gaps within a row. Optical imaging with sensors mounted on Unmanned Aerial Vehicles (UAV) is a cost-effective option for capturing images covering croplands nowadays. However, visual inspection of such images can be a challenging and biased task, specifically for detecting plants and rows on a one-step basis. Thus, developing an architecture capable of simultaneously extracting plant individually and plantation-rows from UAV-images is yet an important demand to support the management of agricultural systems. In this paper, we propose a novel deep learning method based on a Convolutional Neural Network (CNN) that simultaneously detects and geolocates plantation-rows while counting its plants considering highly-dense plantation configurations. The experimental setup was evaluated in (a) a cornfield (Zea mays L.) with different growth stages (i.e. recently planted and mature plants) and in a (b) Citrus orchard (Citrus Sinensis Pera). Both datasets characterize different plant density scenarios, in different locations, with different types of crops, and from different sensors and dates. This scheme was used to prove the robustness of the proposed approach, allowing a broader discussion of the method. A two-branch architecture was implemented in our CNN method, where the information obtained within the plantation-row is updated into the plant detection branch and retro-feed to the row branch; which are then refined by a Multi-Stage Refinement method. In the corn plantation datasets (with both growth phases – young and mature), our approach returned a mean absolute error (MAE) of 6.224 plants per image patch, a mean relative error (MRE) of 0.1038, precision and recall values of 0.856, and 0.905, respectively, and an F-measure equal to 0.876. These results were superior to the results from other deep networks (HRNet, Faster R-CNN, and RetinaNet) evaluated with the same task and dataset. For the plantation-row detection, our approach returned precision, recall, and F-measure scores of 0.913, 0.941, and 0.925, respectively. To test the robustness of our model with a different type of agriculture, we performed the same task in the citrus orchard dataset. It returned an MAE equal to 1.409 citrus-trees per patch, MRE of 0.0615, precision of 0.922, recall of 0.911, and F-measure of 0.965. For the citrus plantation-row detection, our approach resulted in precision, recall, and F-measure scores equal to 0.965, 0.970, and 0.964, respectively. The proposed method achieved state-of-the-art performance for counting and geolocating plants and plant-rows in UAV images from different types of crops. The method proposed here may be applied to future decision-making models and could contribute to the sustainable management of agricultural systems.
\end{abstract}


\section{Introduction}
Advances in both remote sensing and computational vision areas are improving agricultural landscape mapping in the past years (Weiss et al., 2020). This integration is benefiting precision farming in several applications, such as environment control (Hunt and Daughtry, 2018); phenology characterization (Wang et al., 2019a) nutrition evaluation (Delloye et al., 2018; Osco et al., 2019; Osco et al., 2020a), yield-prediction (Chen et al., 2017; Hunt et al., 2019; Jin et al., 2019; Sun et al., 2019); temporal analysis (Zhong et al., 2019), crop-management (Wang et al., 2019b) and others. With the intensification of food demand around the world, farmers are required to increase their efficiency. However, this increase in productivity must come from technological advances and optimization of the production areas instead of their expansion. An accurate estimation of the number of plants in crop fields is important to predict the amount of yield while monitoring growth status (Kitano et al., 2019). Likewise, the detection of plantation-rows is essential since this information can be used by a specialist to evaluate the number of missed plants in each plantation-row and, consequently, the production rate of a crop (Oliveira et al., 2018). These practices can help improve precision farming applications, resulting in better management of the agricultural system.

The visual inspection of plants in agricultural fields is a complicated task because it can be challenging and biased (Leiva et al., 2017). Recently, data obtained from Unmanned Aerial Vehicle (UAV)-based sensors have been used to assist its management (Jiang et al., 2017; Deng et al., 2018). Different sensing systems are used to map plants in high-resolution images, like UAV-based RGB, multi and hyperspectral cameras (Surový et al., 2018; Ozdarici-Ok, 2015; Paoletti et al., 2018), Light Detection and Ranging (LiDAR) (Verma et al., 2016; Hartling et al., 2019), Synthetic Aperture Radar (SAR) (Ndikumana et al., 2018; Ho Tong Minh et al., 2018) and airborne imagery (Li et al., 2016). Sensors such as LiDAR and SAR, although returns a high performance in plant detection (Jakubowski et al., 2013; Tao et al., 2015), are high-priced and difficult to reproduce in low-budget models. To circumvent this, recent studies regarding plant or tree density estimation have implemented RGB-based sensors in their applications (Weinstein et al., 2019; Csillik et al., 2017; Fan et al., 2018; Varela et al., 2018; Ampatzidis and Partel, 2019). The low cost and high market availability associated with them may justify this preference. Furthermore, in the computational vision context, RGB images are enough for straightforward identification tasks such as plant detection (Wu et al., 2019; Hassanein et al., 2019).

The automatic identification of plants is generally divided into two categories: detection and delineation (Özcan et al., 2017). For detection purposes, both plant-size and spatial resolution of the image can be considered enough features (Csillik et al., 2018). Delineation, however, may require information regarding spectral heterogeneity between the scenes’ objects, shadow complexity, and background effects (e.g. soil brightness) (Nevalainen et al., 2017). In the past years, morphological operations and segmentation algorithms like “Watershed”, “Valley Following”, and “Region Growing” were used to count plants in both forested (Larsen et al., 2011) and cultivated areas (Özcan et al., 2015). The aforementioned techniques rely mostly on the spectral divergence between the pixels (plant and non-plant), indicating that a brighter pixel is recognized as the plant, while dark pixels (viewed as shadows) represent their boundary. In such cases, an Excess Greenness Index (ExG) could be applied to individualize the green pixels with high saturation from the background (Varela et al., 2018), or object-based approaches (OBIA) (Hussain et al. 2013) or the use of Fourier-transformations (Jensen, 2015) and Gray Level Co-Occurrence Matrix (GLCM) textural metrics (Huang et al. 2014). Another technique could be the conversion from RGB to grayscale HSV image data (Oliveira et al., 2018). These methodologies obtained interesting results in the last decade. Still, in recent years, more robust and intelligent algorithms are being created and tested in these applications, like Deep Learning (DL)-based models to promote a more generalized approach.

DL is one type of Machine Learning (ML) technique, based on Artificial Neural Networks (ANN), adopting a deep strategy for data representation (Ghamisi et al., 2017; Badrinarayanan et al., 2017), resulting in a large learning capability and improved performance (Ball et al., 2017), in which several components and types of layers constitute a DL architecture (Kamilaris and Prenafeta-Boldú, 2018). The most frequently used architectures in the past years are Unsupervised Pre-Trained Networks (UPN), Recurrent Neural Networks (RNN), and Convolutional Neural Networks (CNN) (Lecun et al., 2015; Khamparia and Singh, 2019). In recent years, the CNN architectures presented great performances for image and pattern recognition, especially in remote sensing approaches (Alshehhi et al., 2017). These approaches can be majorly separated into spectral, spatial, and spectral-spatial information extraction (Ghamisi et al., 2017; Li et al., 2017; Zhang et al., 2017). When considering both spectral and spatial information, the model accuracy can significantly improve (Zhang et al., 2017), which is important since it helps to address the appropriate approach to solve specific problems.

In vegetation detection and delineation, DL models have been used to identify weed in bean and spinach fields (Dian Bah et al., 2018), count palm trees in plantation areas (Djerriri et al., 2018; Li et al., 2017), classify urban-trees species (Santos et al., 2019; Hartling et al., 2019), tree crown prediction in forest areas (Weinstein et al., 2019), counting of rice seedlings (Wu et al., 2019), identification of citrus-tree crowns (Osco et al., 2020b; Csillik et al., 2018), tobacco plant detection (Fan et al., 2018), fir-trees insect-damage detection (Safonova et al., 2019), and others. The models implemented in these studies were mostly derived from the RNN and CNN architectures, some with modified versions of previously published algorithms, while others presenting an entirely new model. Regardless, a recent revision paper indicated that proximally 42\% of the implemented architectures in agricultural studies were based on CNNs, as AlexNet, VGG16, and Inception-ResNet, compared to other deep learning architectures, like Recurrent Neural Networks and Recursive Neural Networks (Kamilaris and Prenafeta-Boldú, 2018).

For the detection of plants, different architectures and modifications were applied in state-of-the-art studies. CNNs adopting the architectures AlexNet (Krizhevsky, 2014) and Inception (v2, v3, and v4) (Ioffe and Szegedy, 2015; Szegedy et al., 2015; Szegedy et al., 2016) had been recently used to count sorghum plants (Ribera et al., 2018). The modification of the mentioned architectures allowed them to estimate the number of plants using regression instead of classification. A modified version of the VGG16 model (Simonyan and Zisserman, 2015) was used to identify tree health status (Sylvain et al., 2019). Likewise, a U-Net (Ronneberger et al., 2015) modification was used to detect palm trees (Freudenberg et al., 2019). A CNN region-based, YOLOv3 (Redmon and Farhadi, 2018), was used to recognize citrus trees and classify its crown (Ampatzidis and Partel, 2019). The YOLOv3 architecture, alongside the RetinaNet (Lin et al., 2017) and Faster R-CNN (Ren et al., 2015) were also used to classify tree species (Santos et al., 2019). Lastly, a modified Deep Convolutional Network (DCN), considering morphological operations and watershed segmentation, was efficiently used to detect tobacco plants (Fan et al., 2018).

The aforementioned methods returned important information regarding the tested approaches in a diverse number of agricultural fields. One type of crop that is still not benefited by these methods is corn. Corn (Zea mays L.) is an important crop and it is largely cultivated in countries like the United States of America, Brazil, China, Canada, and others (Mohanty and Swain, 2019). For the detection of corn-plants in UAV imagery, few studies have been conducted with this computer vision approach (Mochida et al., 2019). An early-season uniformity detection with a decision tree algorithm in an object-detection approach was used to identify corn plants in ultra-high-resolution imagery acquired with UAVs (Varela et al., 2018). Experiments related to drought stress were evaluated through a Deep CNN with images obtained from a stationary station (An et al., 2019). Regarding the use of CNN in UAV imagery, a study applied the U-Net architecture to segment corn-plants from other field objects (Kitano et al., 2019). This study, however, comments on issues that should be addressed in future research, such as earlier growth stages and higher plant density. Another unexplored issue, although not mentioned in these studies, is the detection of plantation-rows.

Another type of agronomic culture that could benefit, both from counting plants and detecting plantation-rows, is Citrus. Citrus tree detection is an important prerequisite for farmers and technicians to estimate yield and even, in some cases, compensate plantation gaps. Recently, studies already discussed the importance of DL in citrus tree counting and area estimation (Osco et al., 2020b; Ampatzidis and Partel, 2019; Csillik et al., 2018). Our previous research (Osco et al., 2020b), conducted with UAV-based multispectral imagery, as well as the others in RGB imagery, returned similar outcomes in this task. However, to the best of our knowledge, a deep learning architecture capable of counting plants and mapping plantation-rows simultaneously for different cultivars is, still, a challenging and unproposed-task. A model with these capabilities can be used as an alternative to the visual interpretation task of crops-fields and could contribute to the sustainable management of agricultural systems. Some crops, such as citrus plants, corn, and many others, have a limited capacity to compensate for missing areas within a row since they cannot occupy those areas or at least lean towards them, and this negatively impacts the yield per unit land area during the harvest season (Primicerio et al., 2017; Varela et al., 2018; Oliveira et al., 2018; Hassanein et al., 2019).

Additionally, performing the counting plant task in high-density areas is a problem even more challenging for both visual inspection and automatic analysis. Few investigations were conducted to solve counting plant tasks in high-density plantations (Osco et al., 2020b; Fan et al., 2018), and in most of these investigations, only one cultivar has been considered. It should also be noted that the detection of plants and plantation-rows consists of an important metric for the assessment of agricultural fields (Primicerio et al., 2017; Hassanein et al., 2019). The number of plants helps farmers and rural technicians to estimate the yield at the end of the crop cycle (Oliveira et al., 2018). This type of assessment, when performed in the early stages of planting, is important for rapid decision making. For corn and other types of cultivars, the decision window is brief, and a rapid detection may help to mitigate or prevent problems with its production. In citrus orchards, the counting of trees is also used to estimate yield and can help farmers to better monitor gaps in their plantation-rows.

In this regard, aiming to contribute to the aforementioned issues, here we present a new deep learning architecture to simultaneously count plants and detect plantation-rows for distinct cultivars from UAV imagery. Our approach is based on a CNN in which its architecture is formed by two processing branches that share information (concatenated) for counting plants and detecting plantation-rows. This allows the refinement of plant detections to the regions where the plantation lines were detected, similarly, the plantation lines learn and adjust to the positions in which the plants are. In our approach, we used RGB imagery to compose a dataset since it is a lower-priced solution than most of the remaining remote sensing systems, being easily replicable in other situations. The framework discussed demonstrates a viable solution with computer vision assistance to count and detect plants and plantation-rows for different types of crops (i.e. corn and citrus) and plantation densities while preserving their geolocation information in UAV-based RGB imagery. The rest of the paper is organized in Section 2, where details of our approach are presented; Section 3, where the results are presented and discussed in Section 4, and; Section 5, which concludes the paper.

\section{Materials and Method}
The framework proposed in this paper was composed of five steps: (1) The acquisition of RGB images from corn and citrus fields with a camera embedded in a UAV platform. (2) The pre-processing of images and their labeling by a specialist in a GIS (Geographical Information System) environment. Here, the specialist defined the plantation-rows using a line-feature and the corn-plants and the citrus-trees with point-features. (3) The data splitting into training, validation, and testing datasets using the hold-out method. (4) The models’ performance evaluation with the detection of the plantation-rows and the number of plants. (5) The error metrics calculated for each task performed by our CNN method.

\subsection{Study Area and Data}
The study was firstly conducted with corn (Zea mays L.) plants in an experimental area undertaken at “Fazenda Escola” at the Federal University of Mato Grosso do Sul, in Campo Grande, MS, Brazil. The evaluated area has approximately 7,435 m², with corn-plants (both young and mature) planted at a 30x50 cm spacing, resulting in 4-to-5 plants per square meter. The plantation-rows consist of two different lengths, lengths, and directions. We considered plants in two growth stages: recently planted (V3) and; mature-stage with cobs. The processed image was labeled by a specialist in the QGIS 3.10 open-source software. Firstly, the plantation-rows were detected using a line feature. Secondly, each line was scanned by the specialist and the corn plants were manually identified by a point feature. Figure 1 shows a high-scale example of the resulting process.

We collected the corn images with a Phantom 4 Advanced (ADV) UAV for two days, using an RGB camera equipped with a 1-inch 20-megapixel CMOS sensor.  The images were obtained with 80\% longitudinal and 60\% lateral overlaps, with a Ground Sample Distance (GSD) of 1.55 cm. The images were processed with Pix4D commercial software. We optimized the interior and exterior parameters of the acquired images and generated the sparse point cloud based on the Structure-From-Motion (SfM) method. We then created the dense point clouds using the Multi-View Stereo (MVS) technique. The UAV flight was approved by the Department of Airspace Control (DECEA), which is responsible for Brazilian airspace.

\begin{figure}[ht]	
\includegraphics[width=\textwidth]{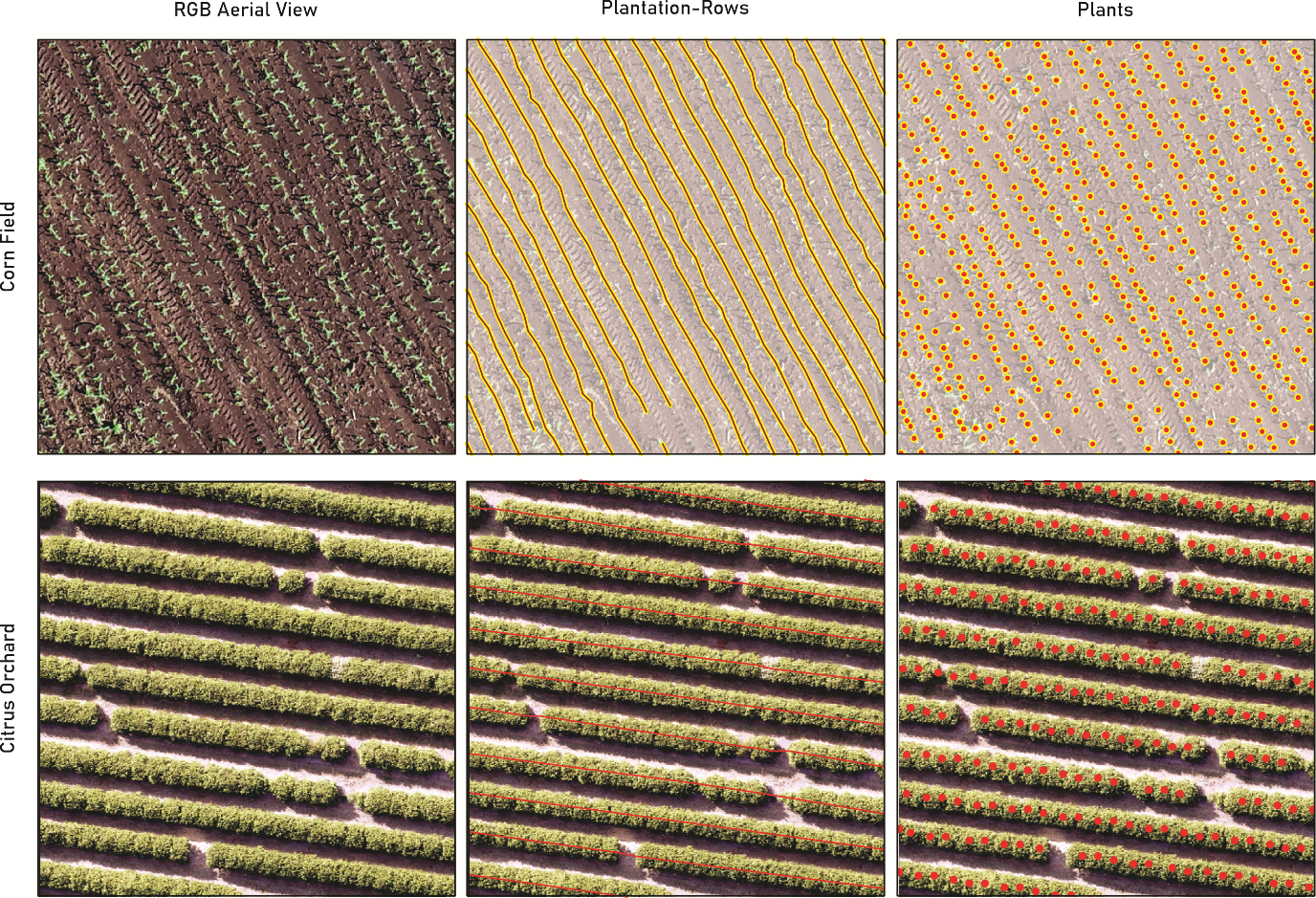}
\caption{High-scale examples of the RGB images used displaying the plantation-rows, corn-plants, and citrus-trees that were manually identified.}
\end{figure}

The second experiment was conducted in a citrus orchard (Citrus Sinensis Pera), located in the countryside at the Boa Esperança do Sul municipality, SP, Brazil. The area is composed of citrus-trees at their maturity phase, in which the spacing in-line was initially around 3 meters from each-others and in the later years, more recent trees were planted at a more approximate area; thus, returning different spacing in-line and densities. The area has approximately 10.000 m². We used an X7 – Spire II UAV embedded with an RGB sensor at 80 meters flight altitude, returning a GSD equal to 2.28 cm. To preprocess these images, we adopted the same strategy mentioned previously using the Pix4DMapper software. Figure 2 illustrates both areas (corn and citrus) with their respective locations and RGB imagery examples.

\begin{figure}[ht]	
\includegraphics[width=\textwidth]{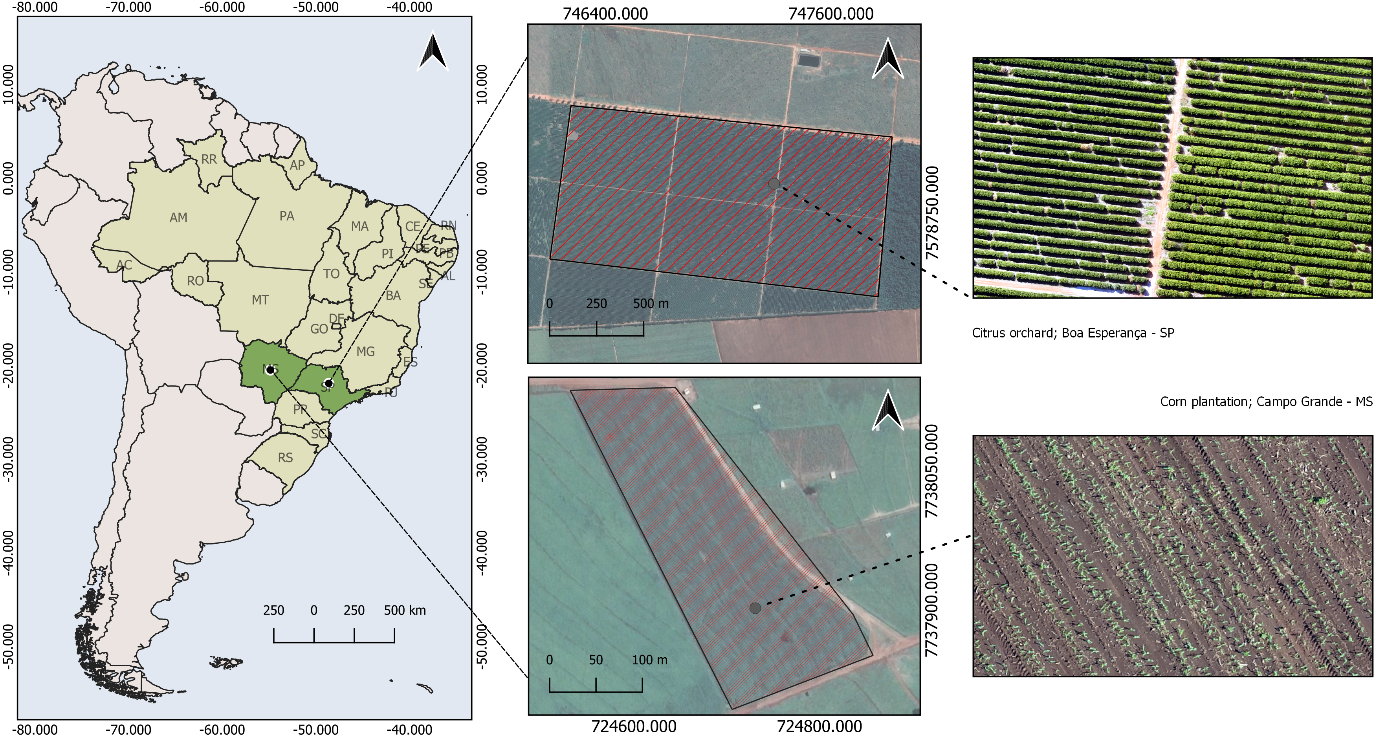}
\caption{Overall visualization of the study area. The cornfield (row-below) is located at Campo Grande, MS, and the citrus orchard (row-above) is in Boa Esperança, SP.}
\end{figure}

\subsection{Convolutional Neural Network}
The UAV images with w ×h pixels were processed using the proposed CNN model to compute the positions of individual plants and plantation-rows. The object counting and geolocation were modeled after a 2D confidence map estimation using the method presented in both Aich and Stavness (2018) and our previous work, Osco et al. (2020b). The confidence map in this case is a 2D representation of the likelihood of an object occurring in each pixel. The Pyramid Pooling Module (PPM) (Zhao et al., 2017) that inserts global and local neighborhood information was included in the model to improve the estimation of the confidence map. A Multi-Stage Module (MSM) prediction that refines the confidence map was used to a more accurate prediction of the center of the objects, similar to our previous work to detect tree species in a national park with hyperspectral imagery (Myioshi et al., 2020). 

Our latest addition to the aforementioned architecture was two detection-branches inside the MSM. This addition was necessary to our model to understand how plantation-rows are displayed in the image and how they are related to the plants’ position, and vice-versa. This construction permitted our deep network to return both lines and point features simultaneously with their respective geolocation information. It should be mentioned, however, that since our method stores the coordinates of the GeoTIFF image format in a separated file, it does not necessarily require this geographic information to perform the detection when analyzing the input image. Regardless, this information is then added later at the final prediction map, where it incorporates the coordinates X, Y from the separated file and creates a geolocated predicted map. The concatenation process, as well as the information exchanged between the two-branches during the multiple stages, allows for the refinement of both confidence maps generated (“object-plant” detection and “line-plantation” detection). With this refinement, the line or point feature is extracted from the centers’ position of the highest peaks on the map.

Figure 3 presents our method for detecting plants and plantation-rows. The method starts by extracting the feature map as shown in Fig. 3(b) from an RGB input image as viewed in Fig. 3(a). The feature map obtains global and local neighborhood information when passing through the PPM (see Fig. 3(c)). The volume is then processed by an MSM, see Fig. 3(d), with T stages, and is refined to detect plants and plantation-rows in two branches. For this, the volume obtained from the PPM module, as shown in Fig. 3(c), is used as an input for the T stages of MSM. Also, both branches share their volumes between each stage of the MSM, obtaining a more precise identification of the plants and plantation-rows. Finally, we obtain the detection of plants as shown in Fig. 3(e) and rows as shown in Fig. 3(f) at the end of the processing of each branch.

The following subsections detail the four main phases of the proposed CNN: Section 2.2.1 presents the generation of the feature map with CNN; Section 2.2.2 shows the feature map enhancement with the PPM module; The refinement of the confidence map by the MSM module is presented in Sections 2.2.3 and 2.2.4; Lastly, Section 2.2.5 shows how we obtain the positions of objects and rows through peaks in the confidence map.

\begin{figure}[ht]	
\includegraphics[width=\textwidth]{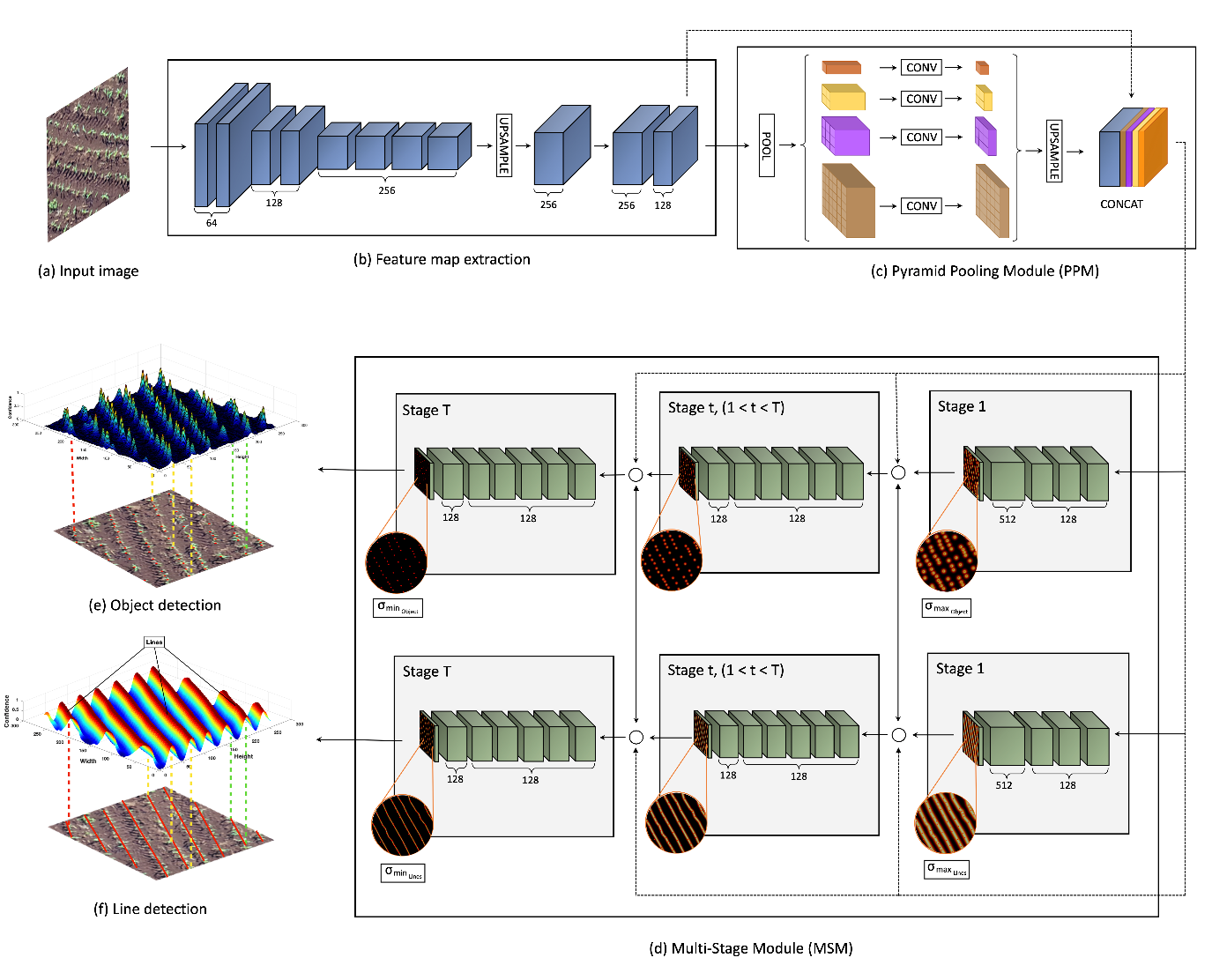}
\caption{Our method proposed for detecting objects and plant-rows: (a) input UAV image, (b) the feature map obtained by CNN, (3) the PPM enhancement module with the feature map as an input, (d) the two detection branches of the MSM, (e) object detection (plants) and (f) line detection (plantation-rows).}
\end{figure}

\subsubsection{Feature Map Extraction}
The feature map was extracted from an RGB input image with 256 × 256 pixels using a CNN based on the VGG19 (Simonyan and Zisserman, 2015) as a feature extractor. The CNN has eight convolutional layers, two maximum pooling layers, and one upsampling layer. The convolutional layers have 64, 128, and 256 filters, all with a size 3 ×3. The two max-pooling layers are inserted after the second and fourth convolutional layers and use a window of 2 × 2. Also, after each convolutional layer, we have rectified linear units (ReLU) function. Finally, the last layer consists of an upsampling layer that delivers an extracted map with 128 × 128 resolution that can describe relevant features from the image.

\subsubsection{Pyramid Pooling Module – PPM}
The global and local image properties allow for the identification of the plants’ position to be more accurate in high-density situations. On the other hand, challenges in identifying plants at different scales and stages of growth are very common in various applications. Thus, our method adopts the global and local context module called PPM (Zhao et al., 2017), which allows it to be scale-invariant and helps the network to deal with multiple sizes of the canopy. The PPM module receives as input the feature map generated in the previous step (see Section 2.2.1) and applies four parallel pooling layers, resulting in four volumes with resolutions of 1 × 1, 2 × 2, 3 x 3, and 6 × 6. The general level, displayed in orange in Figure 4, applies a global maximum pooling that creates a feature map of 1 × 1 to describe the global context of the image. The other levels divide the input map into sub-regions, forming a grouped representation of the image.

The features of each pyramid scale pass through a convolutional layer with 512 filters of size 1 × 1 and are upsampled with the same size as the input map with bilinear interpolation. Lastly, these feature maps are concatenated with the input map to form an improved description of the image. Although this module is commonly proposed for semantic segmentation tasks, it has proven to be a robust method for counting objects according to our experiments, as it includes information at different scales and global context.

\subsubsection{Multi-Stage Module Refinement and Co-Shared Information}
The MSM refinement phase estimates a confidence map from the improved feature map obtained by the PPM. This phase includes two branches of detection with T refinement stages; the first is for plant detection and the second is for plantation-row detection. The first stage contains five convolutional layers and receives as input the improved feature map of the PPM module. The first three layers have 128 filters with 3×3 sizes, the fourth layer has 512 filters with 1 × 1 size, and the last layer is composed of a single filter that corresponds to the confidence map generated by the first stage of each branch, C1\textsubscript{plant} and C1\textsubscript{row}, respectively.

The T-1 final stages refine the positions predicted in the first stage, forming one type of hierarchical learning of the object positions. Because of that, in the stage t, where t=[2,3,…,T], the prediction returned by the previous stage of each branch (C(t-1)\textsubscript{plant}, C(t-1)\textsubscript{row}) and the feature map from the PPM process are concatenated. Later, they are used to produce a refined confidence map for each branch of the stage t (Ct\textsubscript{plant} and C1\textsubscript{row}). These stages have seven convolutional layers, in which: five layers with 128 filters with a 7×7 size; and one layer with 128 filters with a 1 × 1 size. The last layer has a sigmoid activation function so that each pixel represents the probability of the occurrence of an object (values between [0,1]). The remaining layers have a ReLU activation function. 

Sharing volumes between the branches at the end of a stage t allows the learning of the plantation-rows, from the previous stage C(t-1)\textsubscript{row}, to influence the plant prediction in the current stage of Ct\textsubscript{plant}, refining object predictions for regions where plantation-rows have been identified. Similarly, learning the positions of plants from a previous stage C(t-1)\textsubscript{plant} helps the row detection branch to predict more accurate plantation-rows in the current stage, Ct\textsubscript{row}, because they consider the objects' predictions for defining these rows. Also, the use of the improved feature map obtained in the PPM phase at the entrance of each stage allows for multi-scale features, obtained from both the global and local context information, to be incorporated into the refinement process. 

Lastly, to avoid the vanishing gradient problem during the training phase, we adopted loss functions to be applied at the end of each stage of the branches. Each branch (i.e. C(t-1)\textsubscript{plant}, C(t-1)\textsubscript{row}) has its loss function; so while ft\textsubscript{plant} represents the loss function of the plant detection, ft\textsubscript{row} represents the loss function of the plantation-row itself. By the end of it, the general loss functions of each branch are also calculated.

\subsubsection{Confidence Map}
To train our approach, two confidence maps, Ct\textsubscript{plant} and Ct\textsubscript{row} are generated as ground-truth for each stage t by using annotations of the plants and plantation-rows in the image (please see section 2.1). The confidence map is generated by placing a 2D Gaussian kernel at the labeled plants and plantation-rows. Thus, in the object detection (plant), we have high responses in their centers, while in the row detection, we have high responses over the entire extension of the plantation-rows. The Gaussian kernel has a standard deviation ($\sigma$t) that controls the spread of the confidence map peak, as shown in Figure 3.

Our approach uses different values of $\sigma$t for each stage t to refine both the plant and the row predictions during each stage. The $\sigma$1 of the first stage is set to a maximum value ($\sigma$max) while the $\sigma$T of the last stage is set to a minimum value ($\sigma$min). The appropriate values of $\sigma$maxand $\sigma$min were evaluated during the experimental phase of our method. The $\sigma$t for each intermediate stage is equally spaced between [$\sigma$max, $\sigma$min]. Besides, the values of $\sigma$maxand $\sigma$min for each branch are independent and could be different for plant and row detection, generating more accurate maps for each task, depending on the characteristics of the crop and plantation pattern itself.

Figure 5 illustrates two examples of ground-truth confidence maps with three values of $\sigma$t for the V3 corn dataset. The top line of the image shows the confidence map for plant detection while the bottom line presents the confidence map for row detection. Column (a) shows the RGB images and the locations of each plant and row marked by red dots and yellow lines, respectively. The columns (b, c, and d) present the ground-truth confidence maps for $\sigma$t = 0.5, 1.0, and 1.5, respectively; which are responsible for controlling the spread in the top values of our confidence maps. During our experiment, the usage of different $\sigma$ helped us to refine the confidence map, improving its robustness.

\begin{figure}[ht]	
\includegraphics[width=\textwidth]{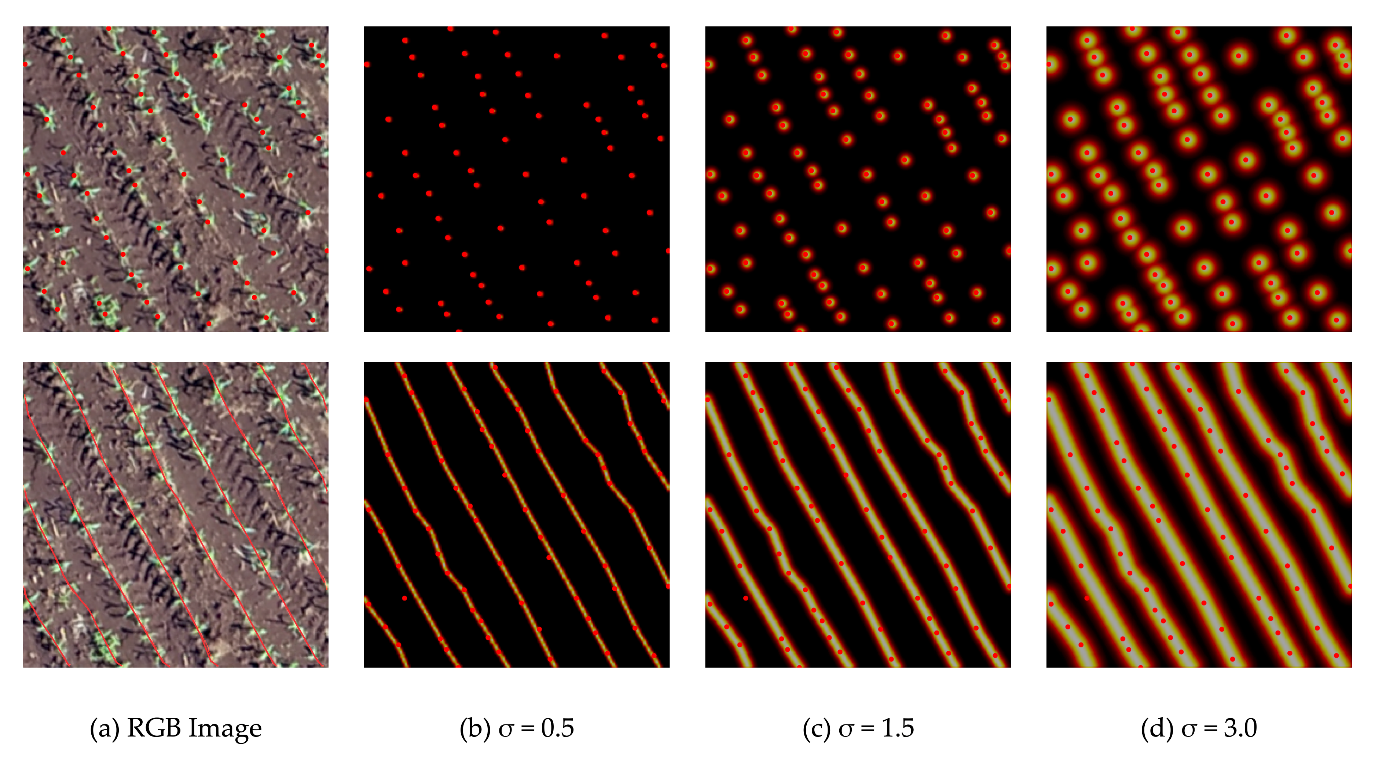}
\caption{Example of an RGB image and its corresponding ground-truth confidence maps for object and row detection with different $\sigma$t values.}
\end{figure}

\subsubsection{Object and Row Localization and Extraction}

The location of plants and plantation-rows is obtained from the last stage of each branch (Ct\textsubscript{plant},Ct\textsubscript{row})  of the MSM module. For the location of the objects, we estimate the peaks (local maximum) of the confidence map by analyzing the 4-pixel neighborhood of each given location of p. Thus, p = (xCt\textsubscript{p}, yCt\textsubscript{p}) is a local maximum when Ct\textsubscript{plant} (p) > Ct\textsubscript{plant} (v) for all the neighbors v, where v is given by (xCt\textsubscript{p} ± 1, yCt\textsubscript{p}) or (xCt\textsubscript{p}, yCt\textsubscript{p} ± 1). An example of the plant location extraction from the confidence map peaks is shown in Figure 5.
 
\begin{figure}[ht]	
\includegraphics[width=\textwidth]{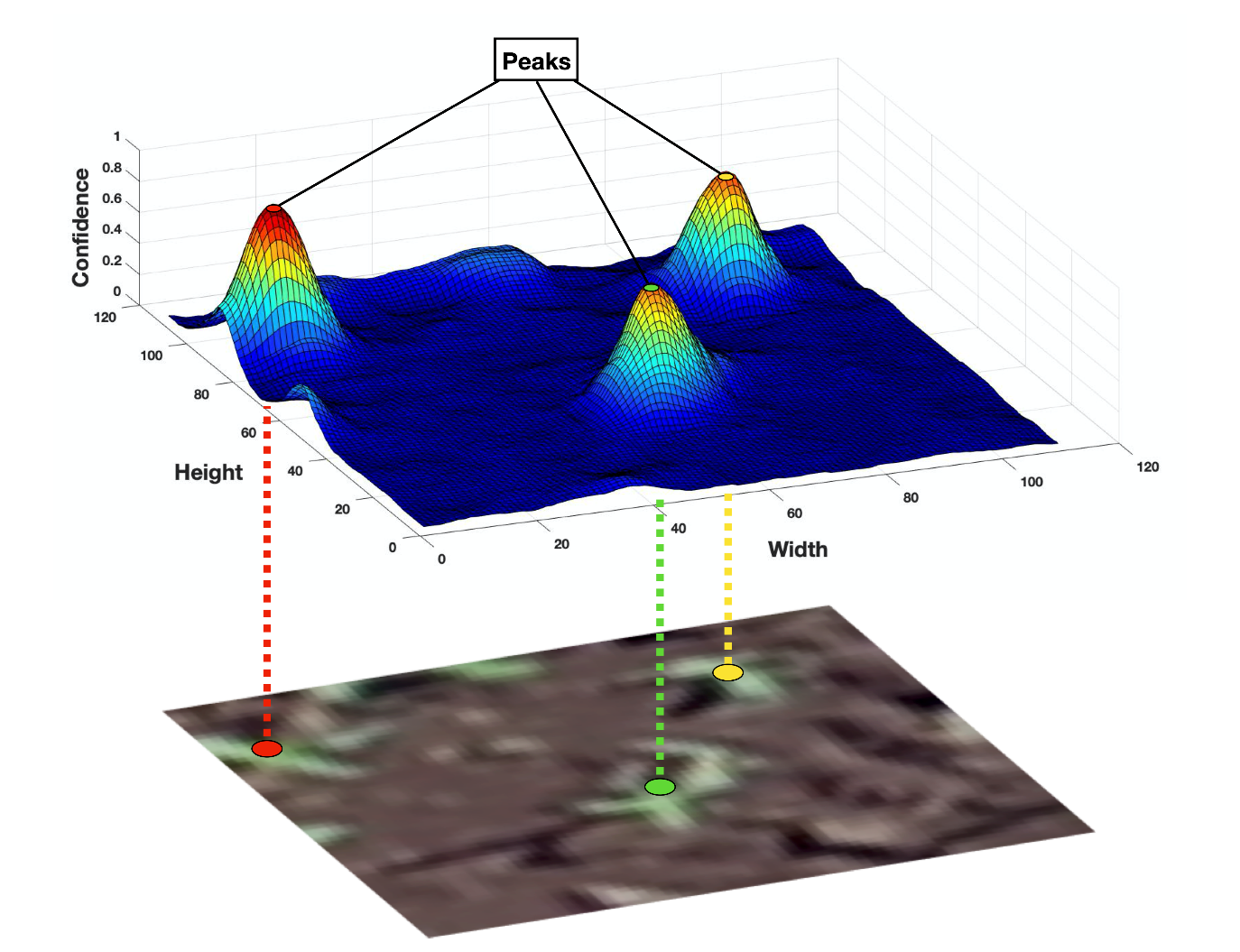}
\caption{Example of the localization of corn plants from a refined confidence map.}
\end{figure}

To avoid noise or low probability of occurrence of the positions of p, a peak in the confidence map is considered as a plant or plantation-row only if CT\textsubscript{(plant-plantation-row)} (p) higher than T. We set a minimum distance to prevent the detection of plants and rows very close to each other. After conducting a preliminary experiment, we used as a minimum 1 pixel and T = 0.35.

To detect rows, we use the skeleton topological algorithm (Zhang and Suen, 1984) on the row confidence map, Ct\textsubscript{row}, to obtain the central activations that represent the plantation-rows. The skeletonization algorithm makes successive passes over the map, removing the activation borders. This process is repeated until there is no border to be removed, and only the skeleton (center-line) of the confidence map remains. Thus, the algorithm delivers a thin version of the shape that is equidistant to its boundaries. Figure 6 shows an example of the confidence map and the skeleton generated by this approach.

\begin{figure}[ht]	
\includegraphics[width=\textwidth]{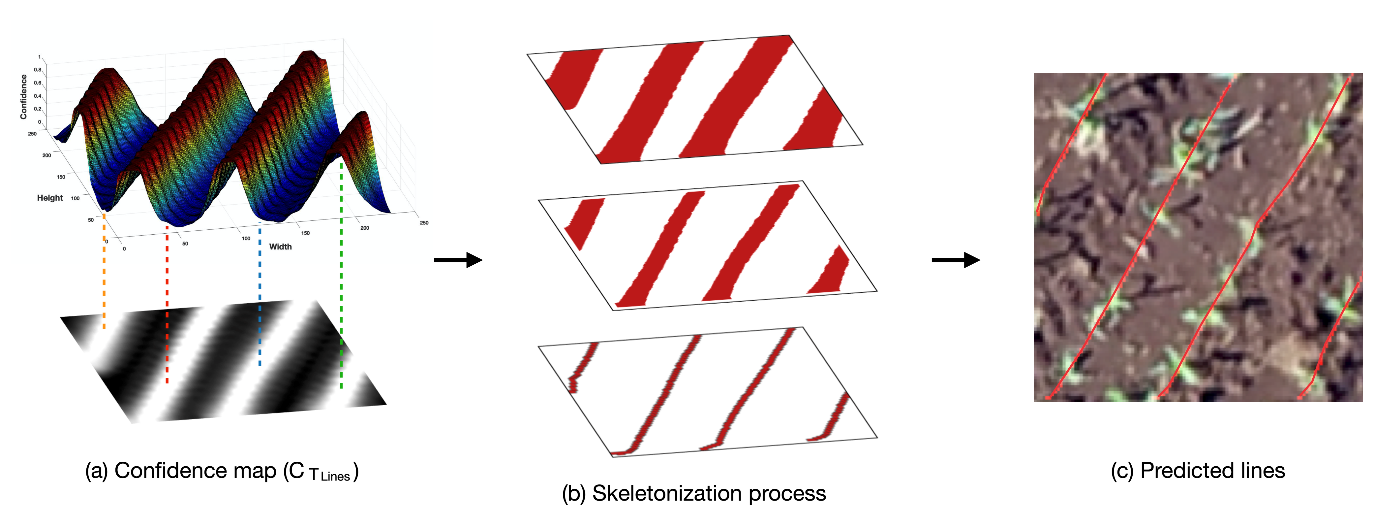}
\caption{Example of skeletonization process on the confidence map. (a) confidence map Ct\textsubscript{row} and its 3D representation, (b) skeletonization process over the confidence map, and (c) predicted rows over the image.}
\end{figure}

\subsection{Experimental Setup}
The collected crop images described in Section 2.1 were split into 564 patches with 256 ×256 pixels without overlapping. The citrus plantation images were split into 635 patches with 256 ×256 pixels without overlapping. The point and line features identified as image-samples were also split between both (corn and citrus) patches. For the corn plantations (V3 and mature), 33,360 plants and 224 plantation-rows were considered in the training phase. As for the citrus-orchard, 14,810 trees and 213 citrus plantation-rows were used in our experiment. As previously stated, this type of characterization with different plant phonologies, sites, and sensor characteristics was essential to ascertain the generalization and robustness of our approach.

The training, validation, and testing sets of each dataset were formed through a random division of the patches in 60\%, 20\%, and 20\%, respectively. This allows the network to not consider the same patch for any of the subsets. For training, we initialize our network with the weights of the first part pre-trained on ImageNet, and apply a stochastic gradient descent (SGD) optimizer with a moment of 0.9. The validation set was used to adjust the learning rate and the number of epochs, reducing the risk of overfitting. After the initial adjustments, the learning rate was set to 0.001 and the number of epochs was set to 100. We also performed additional comparisons with our CNN against state-of-the-art deep neural networks, like HrNet, Faster R-CNN, and RetinaNet. These networks were implemented considering the same dataset characteristics and sampling conditions.

Both our CNN and the other deep networks methods were implemented in Python language on the Ubuntu 18.04 operating system and used the Keras-TensorFlow API. The computer used for training and testing has an Intel (R) Xeon (E) E3-1270 3.80GHz CPU, 64 GB memory, and an NVIDIA Titan V graphics card, that includes a 5120 CUDA (Compute Unified Device Architecture) cores and 12 GB of graphics memory. Finally, to evaluate the performance of our approach, we adopted five regression metrics: mean absolute error (MAE), mean relative error (MRE), the mean square error (MSE), precision (PR), recall (RE), and F1-measure (FM) The regression metrics, MAE, MRE, and MSE, were also adopted in this problem since they estimate the comparison between a given number of labeled corn plants and citrus trees against the predicted positions returned by the network for each image patch. This strategy was also adopted in related work (Osco et al., 2020b) and helped to better explain the estimative and performance of the method. \\

\section{Results}
This section is organized as follows: First, we present the results from an analysis of the parameters to refine the prediction, following the analysis of the $\sigma$ values for the generation of the ground-truth confidence maps of each detection branch. Later, we compare the results from our corn plantation dataset with a baseline method of our CNN as well as other deep learning architectures. Finally, we explore the generalization of our method by evaluating its performance on an entirely different agricultural crop (citrus).

\subsection{Corn Plantation Dataset}

\subsubsection{Parameter Analysis in Plant and Plantation-Row Detection}
Here, we present the results of the proposed method in the validation set for a different number of stages on the corn-crop dataset. These stages are responsible for refining the confidence map. We observed that, by using two stages (T = 2), the proposed method already returned the following results (Table 1). When increasing to T = 6 stages, we obtained the following results in detecting plants: MAE = 7.778, MRE = 0.1360, MSE = 100.132, PR = 0.865, RE = 0.937, and FM = 0.894. These results indicate that the multi-stage refinement affects the plant detection/counting tasks significantly. This is because the confidence map is refined in later stages, increasing the chance of plants to be detected in hard-to-detect regions of the image, associated with a highly-dense plantation. Similarly, we can observe that the performance in predicting plantation-rows is improved with T = 6 stages, reaching PR, RE, and FM of 0.934, 0.983, and 0.956, respectively. Although the result for eight stages (T = 8) is slightly better than T stages = 6 (Table 1), the computational cost significantly increases, not justifying its adoption in later experiments.

\begin{table}[ht]
\begin{center}
\caption{Evaluation of the number of stages T on the validation set using min = 1 and max = 3 for both branches.}
\begin{tabular}{cccccccccc}
\hline
\multirow{2}{*}{Stages   (T)} & \multicolumn{6}{c}{Plant} & \multicolumn{3}{c}{Row} \\
 & MAE & MRE & MSE & PR & RE & FM & PR & RE & FM \\
\hline 
2 & 7.991 & 0.1398 & 102.769 & 0.862 & 0.928 & 0.888 & 0.926 & 0.985 & 0.952 \\
4 & 7.672 & 0.1342 & 98.89 & 0.866 & 0.931 & 0.892 & 0.924 & 0.985 & 0.951 \\
6 & 7.778 & 0.1360 & 100.132 & 0.865 & 0.937 & 0.894 & 0.934 & 0.983 & 0.956 \\
8 & 7.867 & 0.1376 & 100.575 & 0.866 & 0.936 & 0.895 & 0.936 & 0.984 & 0.957 \\
\hline
\end{tabular}
\end{center}
\end{table}

We also evaluated the $\sigma$\textsubscript{min}and $\sigma$\textsubscript{max} responsible for generating the ground-truth confidence maps implemented in the T stages of the MSM phase. In the first stage, the confidence map is generated with $\sigma$\textsubscript{max}, while the last stage uses $\sigma$\textsubscript{min} and the intermediate stages adopt values equally spaced between [$\sigma$\textsubscript{max}, $\sigma$\textsubscript{min}]. The $\sigma$ value concerning the plant influences the predicted location of their locations by the model. A low $\sigma$ provides a confidence map that does not correctly cover the plants’ whole area, while a $\sigma$ too high can include nearby plants in high-density conditions. In both cases, these conditions make it difficult to spatially locate the plants in the image.

To evaluate $\sigma$\textsubscript{min}and $\sigma$\textsubscript{max} we adopted the stages T = 6 that obtained the best results in the previous phase of the experiment, and we use the same value of $\sigma$\textsubscript{min}and $\sigma$\textsubscript{max} in the two detection branches (plant and plantation-row). Table 2 shows the evaluation of $\sigma$\textsubscript{max} in the validation set. The highest result was obtained with $\sigma$\textsubscript{max} = 3, indicating that the confidence map peak, by adopting this value, covers correctly each plant without overlapping nearby plants. On the other hand, as the confidence map is refined by the T stages, and $\sigma$\textsubscript{max} is applied in the first stage, it has a small influence on the final result, evidenced by the results of the other $\sigma$\textsubscript{max}.

Table 3 presents the results of the $\sigma$\textsubscript{min} evaluation with the validation set. In this experiment, we set the $\sigma$\textsubscript{max}= 3 for the two branches and the stages T = 6 because they obtained the best performances in the previous experiments. In this phase, we observe that $\sigma$\textsubscript{min} has a great influence on the method performance since it is used to generate the ground-truth confidence maps at the last stage of the MSM phase. The best result was obtained with $\sigma$\textsubscript{min}= 1.0, indicating a better fit for the plant canopy size. The experiments showed that the best results for counting plants were achieved with $\sigma$\textsubscript{max}= 3 and $\sigma$\textsubscript{min}= 1, delivering F1-measure of 0.894 and MAE of 7.778 plants per patch, respectively.

\begin{table}[ht]
\begin{center}
\caption{Evaluation of the $\sigma$\textsubscript{max}in the validation set. We adopted stages T = 6 and $\sigma$\textsubscript{min}= 1 for both branches.}
\begin{tabular}{cccccccccc}
\hline 
\multirow{2}{*}{$\sigma$\textsubscript{max}} & \multicolumn{6}{c}{Plant} & \multicolumn{3}{c}{Row} \\
 & MAE & MRE & MSE & PR & RE & FM & PR & RE & FM \\
\hline 
2 & 7.867 & 0.1376 & 100.592 & 0.863 & 0.934 & 0.892 & 0.939 & 0.984 & 0.959 \\
3 & 7.778 & 0.1360 & 100.132 & 0.865 & 0.937 & 0.894 & 0.934 & 0.983 & 0.956 \\
4 & 7.734 & 0.1353 & 96.867 & 0.865 & 0.931 & 0.891 & 0.935 & 0.984 & 0.957 \\
\hline 
\end{tabular}
\end{center}
\end{table}

\begin{table}[ht]
\begin{center}
\caption{Evaluation of the $\sigma$\textsubscript{min}in the validation set. We used stages T = 6 and $\sigma$\textsubscript{max}= 3 for both branches.}
\begin{tabular}{cccccccccc}
\hline 
\multirow{2}{*}{$\sigma$\textsubscript{min}} & \multicolumn{6}{c}{Plant} & \multicolumn{3}{c}{Row} \\
 & MAE & MRE & MSE & PR & RE & FM & PR & RE & FM \\
\hline 
0.5 & 21.362 & 0.3737 & 552.070 & 0.949 & 0.597 & 0.723 & 0.944 & 0.984 & 0.962 \\
1 & 7.778 & 0.1360 & 100.132 & 0.865 & 0.937 & 0.894 & 0.934 & 0.983 & 0.956 \\
1.5 & 8.230 & 0.1440 & 111.115 & 0.857 & 0.938 & 0.890 & 0.921 & 0.981 & 0.948 \\
2 & 7.840 & 0.1371 & 98.902 & 0.860 & 0.929 & 0.888 & 0.900 & 0.963 & 0.928 \\
\hline 
\end{tabular}
\end{center}
\end{table}

Regardless of the aforementioned observations, we noticed that the plantation-row detection performance was better with $\sigma$\textsubscript{max}= 2 and $\sigma$\textsubscript{min} = 0.5, reaching PR, R and FM of 0.939, 0.984 and 0.959, respectively, for $\sigma$\textsubscript{max}= 2 (see Table 2) and 0.944, 0.984 and 0.962, respectively, for $\sigma$\textsubscript{min} = 0.5 (see Table 3). Therefore, we evaluated the variation of $\sigma$ between the two branches, which so far were considered the same. To solve this, we defined a $\sigma$ pair for each branch, $\sigma$\textsubscript{min,plant}, $\sigma$\textsubscript{max,plant} and $\sigma$\textsubscript{min,row}, $\sigma$\textsubscript{max,row} for plant and plantation-row detection branches, respectively. For the plant detection branch, we set $\sigma$\textsubscript{min,plant}=1.0 and $\sigma$\textsubscript{max,plant}=3.0 that had the best results in the previous experiments, and we varied the $\sigma$ of the plantation-row detection branch according to Table 4. The experiments showed that the best results were obtained with $\sigma$\textsubscript{min,row}=0.5 and $\sigma$\textsubscript{max,row}=3.0, reaching PR, R, and FM of 0.950, 0.983, and 0.965, respectively.

\begin{table}[ht]
\begin{center}
\caption{Evaluation of $\sigma$ for planting row detection. We adopted the $\sigma$\textsubscript{min, plant} = 1, $\sigma$\textsubscript{max, plant} =3 and stages T = 6.}
\begin{tabular}{ccccccccccc}
\hline 
\multirow{2}{*}{$\sigma$\textsubscript{row,min}} & \multirow{2}{*}{$\sigma$\textsubscript{row,max}} & \multicolumn{6}{c}{Plant} & \multicolumn{3}{c}{Row} \\
 &  & MAE & MRE & MSE & PR & RE & FM & PR & RE & FM \\
\hline 
1 & 3 & 7.778 & 0.1360 & 100.132 & 0.865 & 0.937 & 0.894 & 0.934 & 0.983 & 0.956 \\
0.5 & 3 & 7.672 & 0.1342 & 97.283 & 0.866 & 0.935 & 0.894 & 0.950 & 0.983 & 0.965 \\
0.5 & 2 & 7.831 & 0.1370 & 101.725 & 0.864 & 0.934 & 0.893 & 0.945 & 0.983 & 0.962 \\
\hline 
\end{tabular}
\end{center}
\end{table}

\subsubsection{Plant and Plantation-Rows Extraction}

To analyze the design of the proposed architecture, we compared it with a baseline model that does not include the plantation-row detection branch. The overall best result with the baseline method was obtained with $\sigma$ = 1.0, returning an MAE, MRE, MSE, PR, RE, and FM equal to 6.345, 0.1051, 71.637, 0.870, 0.940, and 0.899, respectively (Table 5). Although the result of $\sigma$ = 0.5 and 2.0 has higher precision value, we observed that in the first case ($\sigma$ = 0.5), has a very low recall value, indicating that the number of false-negative predictions is high. Still, for $\sigma$ = 2.0, we observe that the recall value decreases while the F1-measure maintains the same, indicating a stabilization in the results.

We analyzed the plant and plantation-row detection branches independently to evaluate their performance over the complete method. In both cases, the results of the proposed approach were better, mainly in the precision metric. This indicates that the proposed approach benefits from information exchange between the branches in the MSM phase. Regardless, these results also indicate that both branches can be applied independently, without much loss of performance, as in situations where the plantation-rows are not well defined or in other problems involving a row or centerline detection (e.g., roads).

\begin{table}[ht]
\begin{center}
\caption{Results of the proposed method and its baselines.}
\begin{tabular}{ccccccccccc}
\hline 
\multicolumn{2}{c}{\multirow{2}{*}{Methods}} & \multicolumn{6}{c}{Plant} & \multicolumn{3}{c}{Row} \\
\multicolumn{2}{c}{} & MAE & MRE & MSE & PR & RE & FM & PR & RE & FM \\
\hline
\multicolumn{2}{c}{Baseline ($\sigma$ = 0.5)} & 23.849 & 0.3950 & 738.876 & 0.946 & 0.536 & 0.685 &  &  &  \\
\multicolumn{2}{c}{Baseline ($\sigma$ = 1.0)} & 6.345 & 0.1051 & 71.637 & 0.870 & 0.940 & 0.899 &  &  &  \\
\multicolumn{2}{c}{Baseline ($\sigma$ = 2.0)} & 5.991 & 0.0992 & 68.132 & 8.872 & 0.938 & 0.899 &  &  &  \\
\multicolumn{2}{c}{Proposed Approach} & 5.486 & 0.0908 & 55.982 & 0.878 & 0.934 & 0.901 & 0.950 & 0.979 & 0.963 \\
$\sigma$\textsubscript{min} = 1 & $\sigma$\textsubscript{max} = 3 & 5.778 & 0.0957 & 61.619 & 0.872 & 0.939 & 0.901 &  &  &  \\
$\sigma$\textsubscript{min} = 1 & $\sigma$\textsubscript{max} = 3 &  &  &  &  &  &  & 0.945 & 0.980 & 0.961 \\
\hline 
\end{tabular}
\end{center}
\end{table}

Regarding the different growth periods of the corn plants (V3 and mature), our neural network produced similar performance metrics (Table 6). The V3 detection was slightly better than mature plants’ detection when evaluating the classification metrics (PR, RE, and FM). The V3 detection was also better in plantation-row detection. This could be related to the smaller size of V3 plants, and therefore the occurrence of fewer occlusions in the neural network’s prediction. As for the better plantation-row identification, since the information is exchanged between both CNN’s branches (Fig. 3(d)), where the concatenation from the multi-stage feature extraction is refined with information from both plant and row locations. So, an improved plant detection will often result in an improved plantation row detection. It should be noted that the MAE, MRE, and MSE metrics, which indicate the amount of error produced in each image patch, were slightly worse for the V3 plants than mature plants. It should be highlighted that is hard-to-detect plants located mostly at the edges of the patches (Figures 10 and 11), implying the increase of MAE, MSR, and MSE values.

\begin{table}[ht]
\begin{center}
\caption{Performance of the proposed CNN according to the different growth periods of the corn plantation}
\begin{tabular}{ccccccccccc}
\hline 
\multirow{2}{*}{Trained over:} & \multirow{2}{*}{Tested over:} & \multicolumn{6}{c}{Plant} & \multicolumn{3}{c}{Row} \\
 &  & MAE & MRE & MSE & PR & RE & FM & PR & RE & FM \\
\hline 
V3 + Mature & V3 + Mature & 6.224 & 0.1038 & 66.706 & 0.856 & 0.905 & 0.876 & 0.914 & 0.941 & 0.926 \\
V3 + Mature & V3 & 7.504 & 0.1243 & 92.53 & 0.87 & 0.924 & 0.891 & 0.947 & 0.98 & 0.961 \\
V3 + Mature & Mature & 5.008 & 0.0840 & 42.184 & 0.843 & 0.887 & 0.862 & 0.884 & 0.904 & 0.893 \\
V3 & V3 & 5.486 & 0.0908 & 55.982 & 0.878 & 0.934 & 0.901 & 0.95 & 0.979 & 0.963 \\
Mature & Mature & 4.378 & 0.0734 & 36.848 & 0.872 & 0.872 & 0.87 & 0.887 & 0.918 & 0.901 \\
\hline
\end{tabular}
\end{center}
\end{table}

Figure 7 presents some examples of the results obtained with the proposed method for the V3 corn-crop dataset when adopting the best configuration predefined during the phase of the experiment, $\sigma$\textsubscript{min}= 1.0, $\sigma$\textsubscript{max}= 3.0, $\sigma$\textsubscript{min, row} = 0.5, $\sigma$\textsubscript{max, row} = 3.0, and T = 6. We considered a region around the labeled plant position to analyze qualitatively the prediction within the plant center. The correctly predicted positions are represented by blue dots in the image, and the regions are represented by see-through yellow-circles whose center is the labeled position of the plant. The blue and red dots represent the true and false plant predictions, respectively, while yellow and red circles identify whether the annotated plants were detected or not by the method. The proposed CNN can correctly predict most of the plant’s positions. Even in overlapping plants, the proposed approach can correctly identify the plants' position. We also observed that the method achieves high performance in plantations at different density conditions.

\begin{figure}[ht]	
\includegraphics[width=\textwidth]{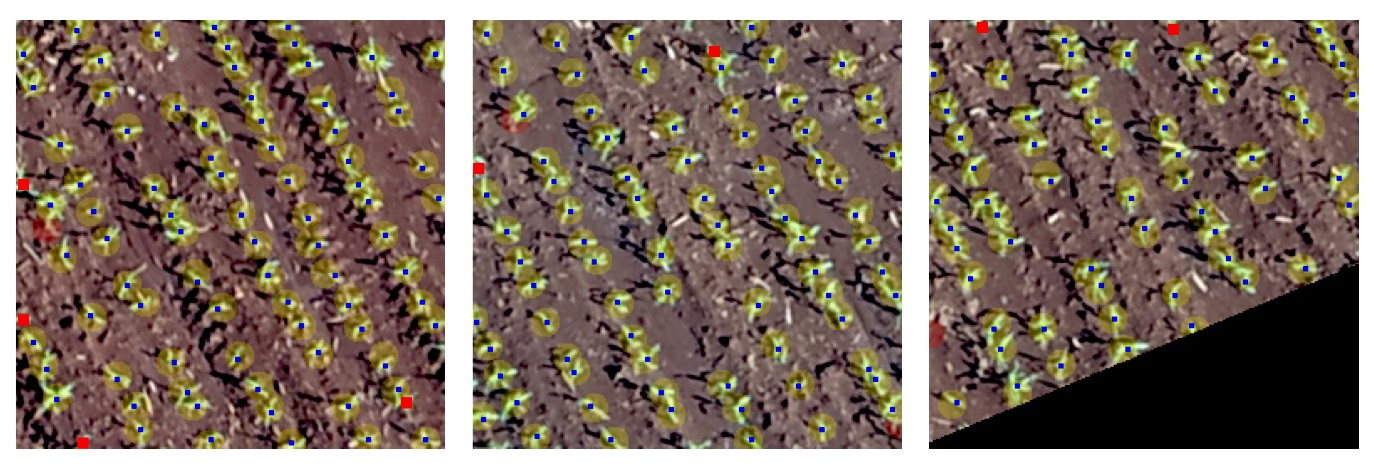}
\caption{Visual results of the proposed method for object detection. Predicted positions are shown by dots while tree-canopies are represented by circles.}
\end{figure}

Although the proposed method is appropriate for most of the corn plant detection and counting tasks, it also faces some challenges (Figure 8). The two main challenges are the plant detection at the borders of the image-patch (see Fig. 8(a)) when most of the plant is occluded, and when we have high-density regions with plants overlapping each other in the plantation-rows (see Fig. 8 (b)). Nonetheless, even in these cases, we observed that our method can correctly predict the position of the majority of the plants. Moreover, the predicted positions of the plants have a high level of accuracy, with most of the predictions (blue-dots) close to the center of the annotations (center of the yellowish circles).

\begin{figure}[ht]	
\includegraphics[width=\textwidth]{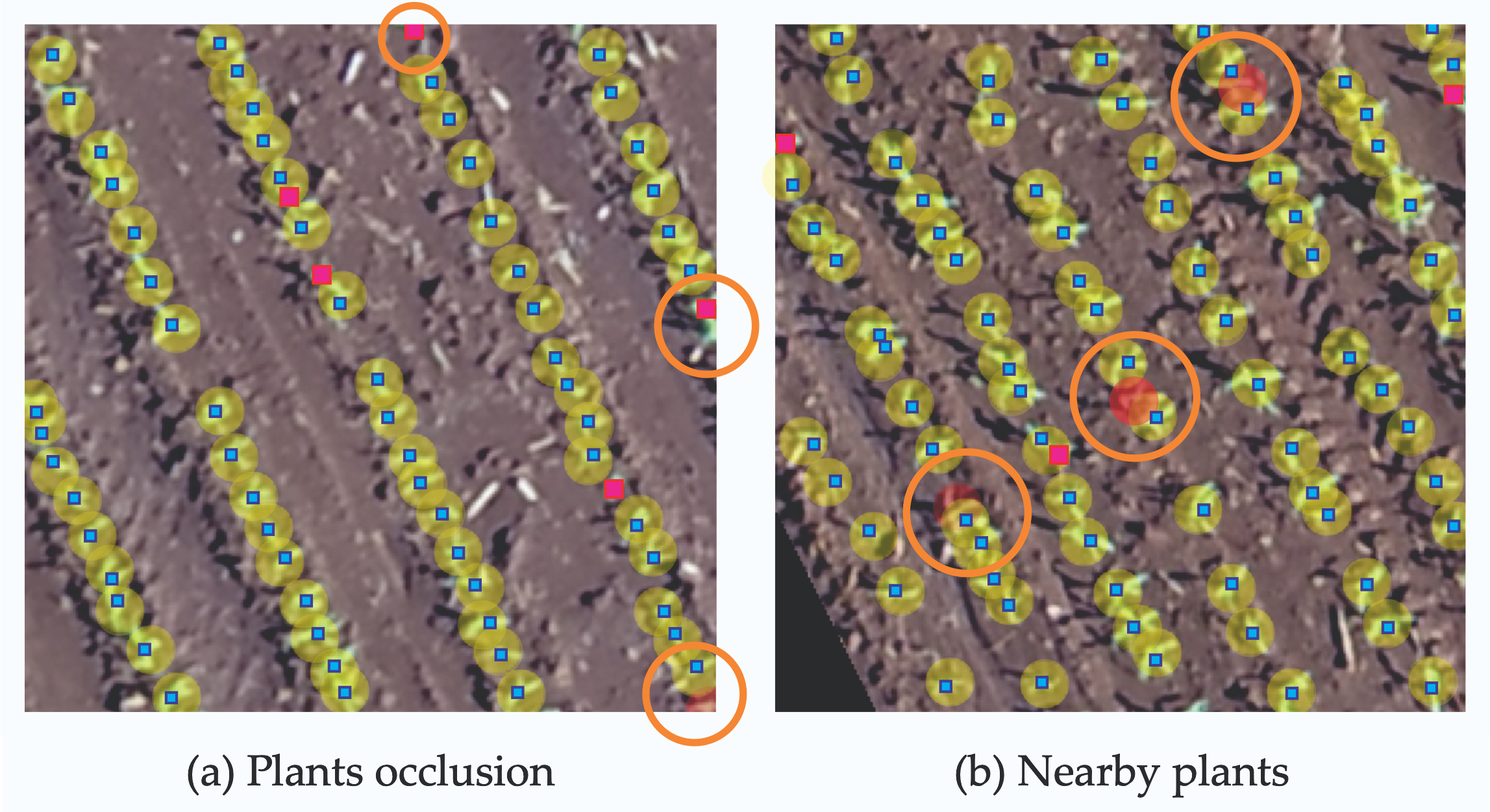}
\caption{Examples of the challenges faced by the proposed method for plant detection. The orange circles show the challenges faced by the method. The blue dots and the yellow circles represent a correct prediction and the tree-canopies of the annotated plants. The red-pink dots and red circles represent the false-positives detections and the missing annotated plants, respectively.}
\end{figure}

Figure 9 shows the performance of the proposed approach in detecting plant-rows. The predicted rows (green) fits with the annotated rows extension (red). The three main challenges in the detection of planting rows are the prediction of rows that correctly adapt the curves of the plantation, the identification of rows with large spacing between plants, and the identification of isolated plants outside the correct plant-row.

As shown in Fig. 9(a), the approach can correctly identify the curves of the plant-rows, highlighted by the orange circles. This happens because when exchanging information between the detection branches, the positions of the detected plants influence the shape of predicted rows. Also, Fig. 9(b) shows that the approach predicted plantation-rows with large spacing between the plants (orange circles). Finally, the identification of outside plants in the plant-rows is a challenge, because they may define plant-rows incorrectly. In this case, the proposed approach can identify them (see Fig. 9(c)) without attributing them to a true plant-row definition.

\begin{figure}[ht]	
\includegraphics[width=\textwidth]{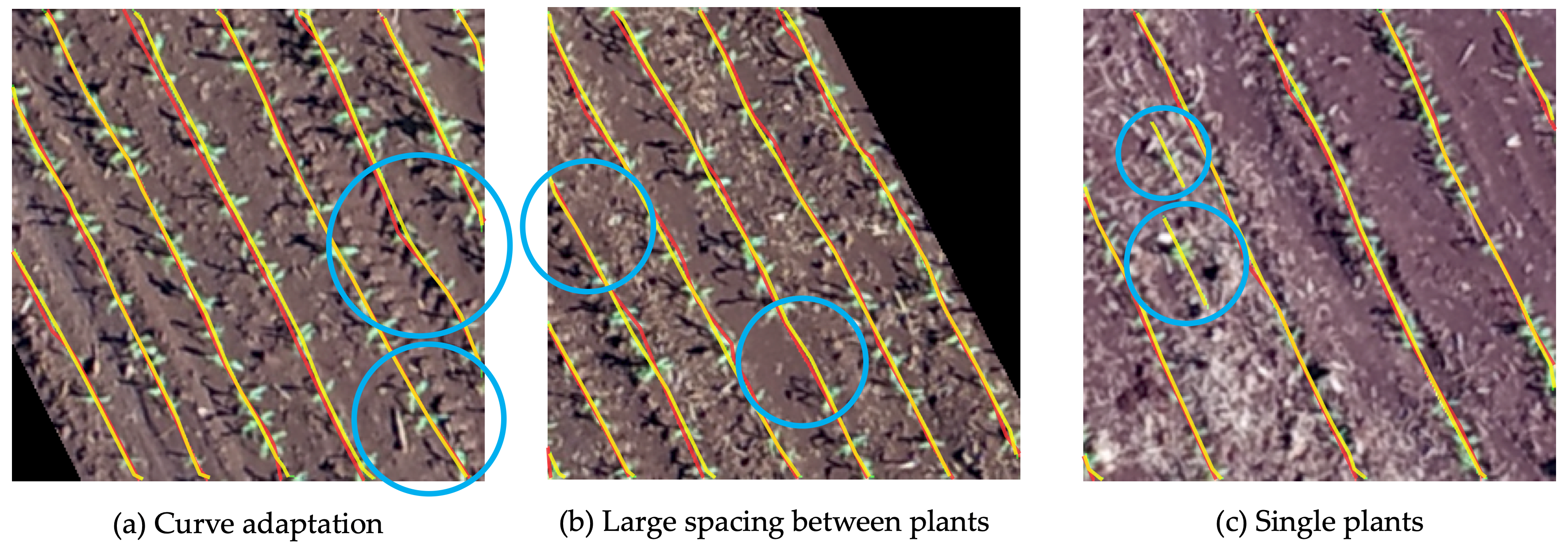}
\caption{Examples of planting row detection by proposed method and its challenges. The blue circles highlight the challenges described. Yellow lines correspond to the lines identified by the network, while red lines under it correspond to the labeled example.}
\end{figure}

\subsubsection{Comparative Results with State-of-the-Art Deep Networks}

Our CNN method obtained better performance when compared to some state-of-the-art object detection methods, like HrNet, Faster R-CNN, and RetinaNet. Table 7 shows the results of this comparison for the MAE, MRE, MSE, PR, RE, and FM metrics. Our approach obtained better values of MAE, MRE, MSE, Precision, and F-Measure when compared with the evaluated methods, reaching important differences in the PR (+9.5\% of HrNet, +12\% of Faster R-CNN, and +16.3\% of RetinaNet) and MAE (-8,665 from HrNet, -11,021 from Faster R-CNN and -14,026 from RetinaNet) metrics. This indicates that the proposed method delivers more accurate detections than the other evaluated deep networks while generating fewer false detections.

Despite the proposed method obtaining slightly lower results for RE, when we analyze the FM, which considers PR and RE, we observed that the approach still obtains better performance with a difference of +3.6\% from HrNet, +5.1\% from Faster R-CNN, and +9\% from RetinaNet. Since the HRNet, Faster R-CNN and RetinaNet baseline methods implemented in this study require bounding-boxes instead of line or point features, we were not able to evaluate its metrics into the plantation-rows properly. Nonetheless, our method achieves high-performance metrics in the detection of plantation-rows, reaching 0.913, 0.941, 0.925 for PR, RE, and FM, respectively. This is an important observation since our method can perform this type of analysis (simultaneously counting and detecting plants and plantation-rows) within a one-step architecture.

To verify the potential of the proposed approach in real-time processing, we compared its performance with the other investigated methods in terms of computational cost. Table 8 shows the average processing time and standard deviation for the detection in the test set. The values of $\sigma$\textsubscript{min,plant}=1, $\sigma$\textsubscript{max,plant}=3, $\sigma$\textsubscript{min,row}=0.5 and $\sigma$\textsubscript{max,row}=3 and T = 6, which obtained the best performance in previous experiments, were used in this evaluation. The results show that the approach can achieve real-time processing, offering image detection in 0.582 seconds with a standard deviation of 0.001. Similarly, the HrNet, Faster R-CNN, and RetinaNet methods obtained an average detection time and standard deviation of 0.070, 0.053, 0.050, and 0.009, 0.010, 0.010, respectively. The processing time was calculated considering the workstation described previously (see section 2.3).

\begin{table}[ht]
\begin{center}
\caption{Results of the proposed method and the object detection methods HRNet, Faster R-CNN, and RetinaNet for the corn plantation (V3 and matures) datasets.}
\begin{tabular}{cccccccccc}
\hline 
\multirow{2}{*}{Methods} & \multicolumn{6}{c}{Plant} & \multicolumn{3}{c}{Row} \\
 & MAE & MRE & MSE & PR & RE & FM & PR & RE & FM \\
\hline 
HRNet & 14.879 & 0.2481 & 319.258 & 0.761 & 0.955 & 0.840 &  &  &  \\
Faster R-CNN & 17.245 & 0.2876 & 392.754 & 0.736 & 0.952 & 0.825 &  &  &  \\
RetinaNet & 20.250 & 0.3377 & 558.025 & 0.693 & 0.940 & 0.786 &  &  &  \\
Proposed Approach & 6.224 & 0.1038 & 66.706 & 0.856 & 0.905 & 0.876 & 0.913 & 0.941 & 0.925 \\
\hline 
\end{tabular}
\end{center}
\end{table}

\begin{table}[ht]
\begin{center}
\caption{Processing time evaluation of the compared approaches.}
\begin{tabular}{ccc}
\hline 
Method & Average   Time (s) & Standard   deviation \\
\hline 
HRNet & 0.070949 & 0.009826 \\
Faster R-CNN & 0.053500 & 0.010663 \\
RetinaNet & 0.050697 & 0.010225 \\
Proposed Approach & 0.582698 & 0.001175 \\
\hline 
\end{tabular}
\end{center}
\end{table}

Figure 10 shows the visual comparison of object detection methods applied here to the corn plants dataset in the advanced growth stage (mature stage with cobs). The proposed approach also returns more accurate detection than the compared methods. The orange circles show the areas where our method stands out in comparison to the other CNNs. We can observe that the evaluated methods generate false detections in denser planting regions. Regardless, in some cases, they detect corn plants outside the planting lines or in empty regions. On the other hand, even in challenging situations highlighted by the blue circles, we notice that our approach delivers more accurate detection than the evaluated methods, with a lower number of false detections (represented by the red dots).
Figure 11 shows the performance of the four methods in detecting corn plants in their initial growth stage (V3). In the usual detection situations (highlighted by orange circles), where there is no overlap and occlusion, our method outperforms the compared methods. Also, the compared methods (Figure 12 (b), (c), and (d)) fail more at the limits of the planting lines and generate nearby false detections. In challenging detections (highlighted by blue circles), our approach delivers more accurate detections than the evaluated methods.

\begin{figure}[ht]	
\includegraphics[width=\textwidth]{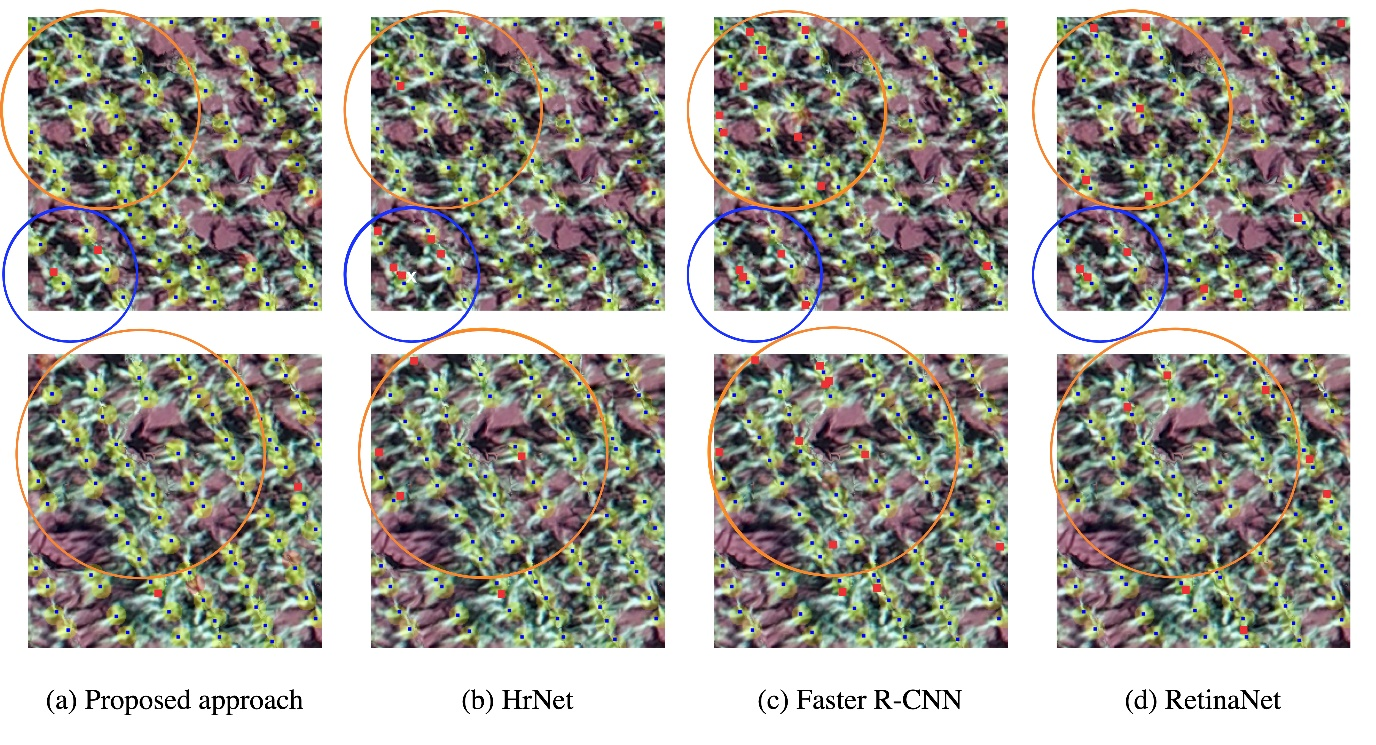}
\caption{Comparison of the object detection methods HRNet, Faster R-CNN, and RetinaNet in the corn plants dataset with a higher growth stage (mature with cobs). The orange and blue circles highlight usual and challenging detections, respectively.}
\end{figure}

\begin{figure}[ht]	
\includegraphics[width=\textwidth]{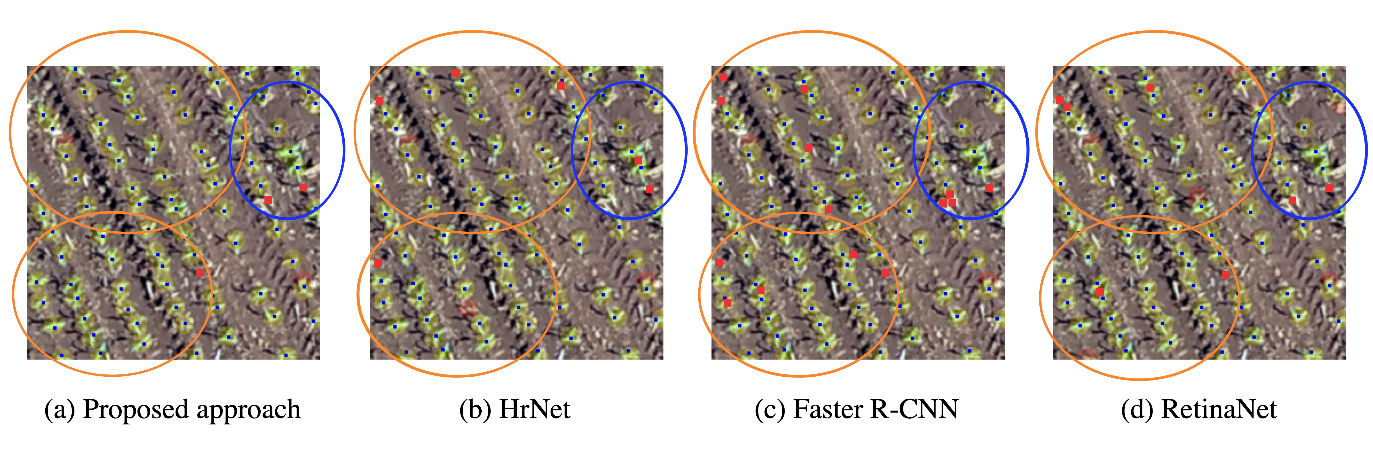}
\caption{Comparison of the object detection methods: HRNet, Faster R-CNN, and RetinaNet in the corn plants dataset with an earlier growth stage (V3). The orange and blue circles highlight usual and challenging detections, respectively.}
\end{figure}

\subsection{Experiments in the Citrus Plantation Dataset}
The parameters used for the citrus plants are the same as the ones adopted for the corn plans, $\sigma$\textsubscript{max,plant}=3, $\sigma$\textsubscript{min,row}=0.5 and $\sigma$\textsubscript{max,row}=3 and T = 6), except for the maximum pixel distance between the prediction and the ground-truth annotation (15 pixels for the citrus radius, while in corn we used 8 pixels), which is justified due to the difference in size between the corn plants and citrus trees canopies areas. The achieved results show that our approach can be generalized into different types of plantations with minimal adjustments to the model (Table 9). Also, even in high-density plantations such as citrus, our method maintains high performance, delivering highly accurate predictions. This shows that the MSM module helps not only by learning information related to plant growth in the multiple stages but also by learning other plantation types with more challenging canopies (high-density).

\begin{table}[ht]
\begin{center}
\caption{Results of the proposed method for the citrus orchard dataset.}
\begin{tabular}{cccccccccc}
\hline 
\multirow{2}{*}{Metrics} & \multicolumn{6}{c}{Plant} & \multicolumn{3}{c}{Row} \\
 & MAE & MRE & MSE & PR & RE & FM & PR & RE & FM \\
\hline 
Proposed Approach & 1.409 & 0.0615 & 3.724 & 0.922 & 0.905 & 0.911 & 0.965 & 0.970 & 0.964 \\
\hline 
\end{tabular}
\end{center}
\end{table}

Figure 12 shows the performance of the proposed method on the citrus dataset. Different from the corn dataset, these plants have a much more complex delimitation, with a large canopy and a lot of overlap between more than one plant. However, following the quantitative results, we found that the proposed approach detections are accurate, delivering centralized detections to the plantation-rows. In the plant detection task (Fig. 12, top row) the blue-dots and yellow-circles represent the correct detections and the tree-canopies of the labeled plants, annotated by a specialist. In the plantation-rows detection (Fig. 12, bottom row) the red-lines represent the annotations by the specialist and the green lines represent the detections by the proposed method. Here, our method overcomes different challenges such as the highly vegetated cover area, that generates overlapping trees canopies (Fig. 12 (a)), the detection with plants and plantation-rows occlusion (Fig. 12 (b)), and the detection of single plants in the limits of the plantation-rows (Fig. 12 (c)).

\begin{figure}[ht]	
\includegraphics[width=\textwidth]{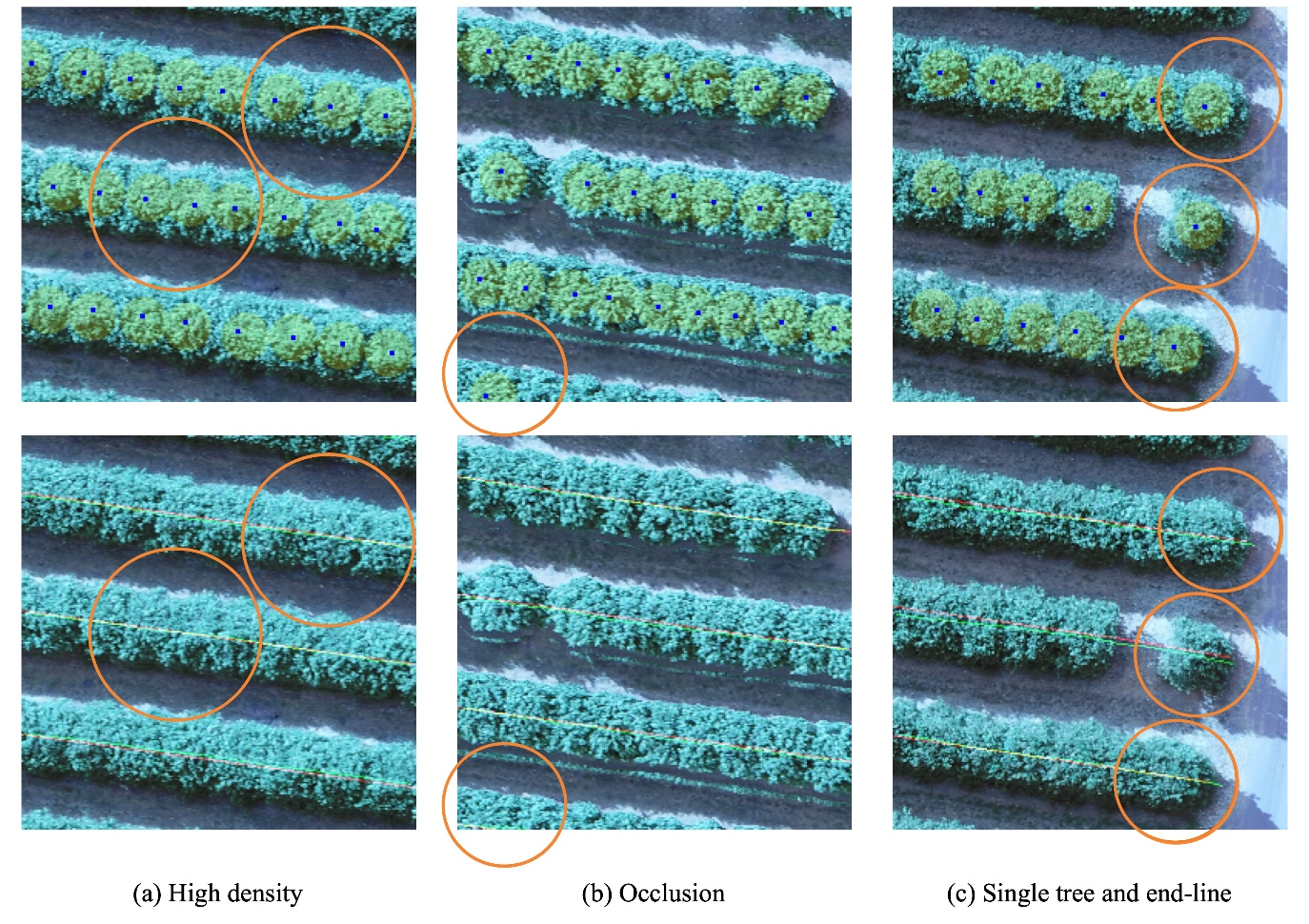}
\caption{Examples of plant and plantation-rows detections in the citrus dataset. Plant and plantation-rows detections are shown in the top and bottom row of the image, respectively. The blue-dots and the yellow-circles represent a correct prediction and the tree-canopies of the labeled plants. The red and green lines represent the annotated and detected plantation-rows. Orange circles highlight the challenges overcome by the approach in each scene.}
\end{figure}

\section{Discussion}
The contribution of our study is to demonstrate a feasible alternative to correctly predict the actual number of plants while simultaneously detecting plantation-rows in UAV-based RGB imagery. Upon our evaluation with different sets, the method presented here for estimating plants and plantation-rows may be replicated in other crops, not being only restricted to the ones presented here. Another important contribution of this method is the detection of high-dense canopies plantations. The usage of a confidence map boosted by the refinement between the two architecture branches helped our network to better detect both overlapped plants and individual plants with high accuracy. Also, another contribution of our method is the plantation-rows refinement, which can help farms to correct problems that occurred during the seedling process at early stages in corn, or compensate for plantation gaps in its area. This form of contribution (plantation-row detection) is also viewed as feasible by similar studies (Primicerio et al., 2018; Fan et al., 2018; Kitano et al., 2019; Salamí et al., 2019; Ampatzidis and Partel, 2019). 

In our corn-field experimental dataset, up to 4 or 5 plants were estimated per square meter (m²) through the photo-interpretation point feature extraction, and our method was capable of detecting these plants with high accuracy levels (F-Measure of 0.876 and MAE of 6.224). An important issue is that the proposed network only needs points and line features, as opposed to the bounding-box type of label needed for the other neural networks. This decreases the time and effort required for the annotation task. Also, our CNN method achieved interesting results with RGB imagery. This tendency to migrate to RGB sensors achieved important outcomes, enough to be a reliable alternative from more expensive equipment such as LiDAR or multi and hyperspectral sensors. RGB images have a relatively low cost to obtain since most conventional UAVs come equipped with them. The tendency to adopt RGB images is not unique to our study (Chen et al., 2017; Wu et al., 2019; Safonova et al., 2019; Salamí et al. 2019). Even so, we intend to implement other types of sensors and methods into our research to take advantage of the proposed method. We believe that this could assist our network in discriminable count plants and detect plantation-rows in other types of environments. Regardless, both datasets evaluated here (corn and citrus) can be considered a high challenge, as they not only are highly-dense types of plantations, with different canopies sizes and pixel-types, but they also represent different plant characteristics.

In cornfields, studies contributed to plant detection in remote sensing imagery. In early-season detection, a decision tree algorithm was able to detect with 0.93 accuracy corn plants with two-to-three leaves (Varela et al., 2018). The closest study to our approach comes from Kitano et al. (2019). Their method used a U-Net architecture modification capable of counting corn plants in different growth stages and flying heights. The best overall result had a 2.6\% residual percentage, while the worst scenario resulted in a 53.3\% residual percentage. Another paper (Gnädinger and Schmidhalter, 2019) performed a digital counting in maize cultivars in aerial images from UAVs, achieving correlations up to 0.89 (R²) and demonstrating the feasibility of their method. One type of deep neural network (DeepSeedling) was also recently developed based upon a Faster RCNN model and achieved F1 scores of 0.727 (at IOUall) and 0.969 (at IOU0.5). In this regard, another research also implemented the Faster RCNN model modification to detect maize in terrestrial imagery, achieving similar scores (Quan et al., 2019). In this manner, we also adopted the Faster RCNN into our models’ comparison; but it returned an inferior F-Measure than the proposed CNN for the same dataset. Our approach, however, presents a lower RE value, which may be associated with the presence of False-Negatives in our method’s prediction. But, since it resulted in higher PR values than the others, the harmonic metric (F) was overall better. Lastly, a study that aimed at segmenting individually maize plants with also the Faster R-CNN and LiDAR imagery returned an accurate model also identifying the measured height with impressive results (R² higher than 0.9). In this manner, another related work developed a method based on an integrated skeleton extraction and pruning approach, also resulting in interesting outcomes (Zhou et al., 2018).

To show the generalization capability of our model, we also performed additional experiments in a citrus orchard. Citrus-tree crowns, in contrast to the corn plants, are planted in a much denser condition. This highly-density system is commonly implemented in multiple regions around Brazil and other countries, as it helps farmers to maintain a high production set and still not expand their farmlands to new areas. Regardless, this type of system is imposing a new difficulty for the deep learning object detection approach, and our CNN method, based upon a confidence map extraction, appears to be suitable to deal with this condition. As not only citrus orchards possess the mentioned characteristic, other types of crops are also planted at much denser states. Thus, it is important to consider this dataset as an interesting challenge to the proposed approach. Regardless, our method was able to return high accuracies even when considering these characteristics (Table 9).

Still, regarding citrus orchards, our previous research (Osco et al. 2020b), with a simpler conception of a CNN, was able to predict citrus-trees in a different dataset composed only from multispectral imagery; returning an MAE equal to 2.28 trees and an F-measure score of 0.950. Other studies conducted in RGB based imagery (Ampatzidis and Partel, 2019; Csillik et al., 2018) were also able to perform well with CNN based architectures (YOLOv3 and a basic CNN), returning classification metrics above 90.0\%. Still, since they evaluated datasets with different characteristics than ours, it is difficult to perform such comparisons. Besides, plantation-row identification is also a not commonly found task in the literature. Recently, a novel deep neural network (CRowNet) was proposed for this task alone (Bah et al., 2019), where the authors showed their approach to detect rows in different crops. 

Another issue that could be further explored with our network is the addition of datasets with a high variety of weeds and other related problems into our plantation-rows. In our area, weeds were not mainly a concern. However, as we concatenated these branches at the end of each stage, the network can handle the target plant detections with more precision, since it determines that, for a plant to exist within a certain position, it needs to share information with its plantation-line. This reduces the possibility of the neural network detecting additional objects with similar spectral information (like weeds) aside from the plantation pattern. Besides, the given patterns of weeds are spatially different from the target plants, so this information is also considered by the CNN when learning the labeled data. Still, the performance of a similar network was evaluated in a previous study with a much higher weed density plantation (Osco et al., 2020b), and returned a similar performance. In this manner, another study demonstrated how an advanced encoder-decoder network was able to outperform other approaches into automatically detecting both crop-rows and weed within the lines (Adhikari et al., 2019). This type of approach could be also considered in our network. Regardless, our neural network, by accurately identifying plants and plantation-rows in both dense and sparse environments, with a one-step type of approach, may still help future research to solve part of this generalization problem when considering both plant and plantation-row identification in one single step.

As previously stated, other crops may benefit from the approach presented here. Whether in counting plants, as well as detecting existing plantation-rows. Since it uses a confidence map where it is calculated the probability that the plant or tree will occur at each pixel, it differs from common object detection deep learning methods that need labeling rectangles to detect a target. In densified plantations, this characteristic of a common object detection deep neural network can be problematic since overlapping plants may reduce the performance of the used model (Ampatzidis and Partel, 2019). Regardless, the highest benefit from the method presented here is the incorporation of a two-branched architecture to deal with plant and plantation-row detection simultaneously on a one-step basis. Since the multiple stage refinement branches are concatenated with each other, both detection approaches (plant and plantation-row) are beneficiated from the knowledge extracted in the other counterpart. In short, as the networks update the branch of the plantation-row with information from the plant branch, the plant branch predictions are refined with information from the plantation-row branch; and vice-versa. Although we also proved, by evaluating each branch prediction individually during our experimental phase in our results section, the overall accuracy of both predictions was higher when considering this strategy.

The proposed CNN is modified to return a prediction map instead of classification. For this reason, the R2, MAE, and MSE metrics differ from the commonly found metrics used in this situation. However, studies that approached plant detection and counting as a classification problem obtained 93.3\% accuracy for rice seedlings using a combined deep network (Wu et al., 2019), 96.2\% for citrus-tree detection (Csillik et al., 2018), and more than 96.0\% to count oil palm trees (Li et al., 2019). In a study that addressed this problem as a prediction, the authors evaluated palm-trees and stated that their AlexNet CNN architecture returned 0.99 R² predictions, with 2.6\% to 9.2\% relative error, depending on the evaluated dataset (Djerriri et al., 2018). Thus, it is evident that the CNN proposed in our study achieved results similar to or better than those existing in the current literature. It should also be considered the adverse situations evaluated here; as it presented more plants per area than the usual. Also, none of these studies implemented a plantation-row detection in their methods; which is another differential of our approach. Although many object detection deep networks can be used to detect plants and rows, they require several modifications to simultaneously perform both tasks. As mentioned, our approach uses a two-branched architecture, and one branch benefits from the other). This interaction between both branches is an important feature and does come in handy when both problems are being considered.

Currently, applications involving UAVs, RGB sensors, and deep learning models are contributing to addressing the aforementioned issues here discussed (Weinstein et al., 2019; Csillik et al., 2017; Fan et al., 2018; Ampatzidis and Partel, 2019). But a disadvantage of deep learning, in general, is the need to label thousands of plants for the training process, as well as a high-end computer to process these data (Goldman et al., 2019). One of the reasons for reducing accuracy in our approach was plant-occlusion and high-proximity between plants (Figure 8) and the occurrence of single-plants outside the plantation-rows, an uncommon spacing between plants, and some “curves” in the plantation-rows (Figure 9). Another issue is related to the consumed time of our approach concerning other methods (Table 8). This occurred mainly because our neural network computes both plants and plantation-lines, while the other compared bounding-box methods only account for plant detection.  Also, on this matter, the usage of object detectors architectures is an alternative to to the adoption of confidence maps and point-labeling over the input image, but the annotation of the plants with bounding boxes, as required by neural networks such as Faster-RCNNN, is more laborious when compared to a single point annotation, as required by our approach. Besides, object detectors have higher difficulties in detecting plants in dense areas as shown in previous works (Osco et al., 2020). Also, the method presented here can be fed with other data sources, as demonstrated with the variated dataset tested, and will require less prior information than before. In this way, where new information is incorporated into the method, more accuracy and new learning patterns can be expected to be achieved.

\section{Conclusion}
This paper introduces a CNN approach to simultaneously detect plants and plantation-rows in different datasets (cornfields and citrus orchards), derived from RGB imagery acquired with a UAV-based remote system. The presented method is a feasible alternative from visual inspection and should assist in precision farming practices. It proved highly accurate results, achieving a mean absolute error (MAE) of 6.224 plants per image patch, MRE of 0.1038, precision and recall values of 0.856 and 0.905, respectively, and an F-measure equal to 0.876. These results were superior to the results from other deep networks (HRNet, Faster R-CNN, and RetinaNet) evaluated with the same task and dataset. For the plantation-row detection, our approach returned precision, recall, and F-measure scores of 0.913, 0.941, and 0.925, respectively. To test the robustness of our model with a different type of crop, we performed the same task in the citrus orchard dataset. It returned an MAE equal to 1.409 citrus-trees per image patch, MRE of 0.0615, precision of 0.922, recall of 0.911, and F-measure of 0.965. For the citrus plantation-row detection, our approach resulted in precision, recall, and F-measure scores equal to 0.965, 0.970, and 0.964, respectively. The proposed method also has a reasonable cost alternative, since it uses an RGB-based sensor.

Another contribution of our CNN approach is that, by applying a two-branched architecture and enabling information to be exchanged between them, our approach can benefit from the results of one detection to the other. Besides, instead of using a common bounding-box object-detection approach, it estimates a confidence map to detect individual plants. This presents an advantage when evaluating high-density plantations, since it does not rely on target boundaries, but it uses the probability of a unique pixel being recognized as the plant. This architecture is also beneficiated from the two branches’ approach as their refinement is linked to the information exchanged between them. As some plants are naturally limited to compensate for missing areas, we recommend the method here to count and detect plantation-rows simultaneously, in a one-step architecture, since it helps to estimate plantation patterns and errors. We intend to implement new features in our method to overcome different challenges regarding plantation patterns in the future. Even so, we trust that, in the current state, the method provides an enhancement in decision-making tasks while contributing to the more sustainable management of agricultural areas by remote sensing systems. We hope that the approach presented here may assist in research regarding remote sensing technologies and precision farming applications.

\section*{Funding}
The authors are funded by EMBRAPA (p: 23700.19/0192-9-1), National Council for Scientific and Technological Development (CNPq) (grant number 303559/2019-5, 433783/2018-4, 310517/2020-6, 313887/2018-7, and 304173/2016-9), CAPES-Print (p: 88881.311850/2018-01), and Fundect (p: 59/300.066/2015 and 59/300.095/2015). V. Liesenberg is supported by FAPESC (2017TR1762).

\section*{Acknowledgments}
The authors acknowledge the support of UFMS (Federal University of Mato Grosso do Sul) and CAPES (Finance code 001), the donation of a Titan X and a Titan V by ©NVIDIA, the EMBRAPA (Brazilian Agricultural Research Corporation) for providing additional imagery datasets, and CAPES (Coordination for the Improvement of Higher Education Personnel - Finance code 001);.

\section*{Conflicts of Interest}
The authors declare that they have no known competing financial interests or personal relationships that could have appeared to influence the work reported in this paper.

\section*{References}
Adhikari, S.P., Yang, H., Kim, H., 2019. Learning Semantic Graphics Using Convolutional Encoder–Decoder Network for Autonomous Weeding in Paddy. Front. Plant Sci. 10, 1–12. https://doi.org/10.3389/fpls.2019.01404

Aich, S., and Stavness, I., 2018. Improving object counting with heatmap regulation. arXiv preprint arXiv:1803.05494.

Alshehhi, R., Marpu, P. R., Woon, W. L., Mura, M. D., 2017. Simultaneous extraction of roads and buildings in remote sensing imagery with convolutional neural networks. ISPRS J. Photogramm. Remote Sens.,130, 139–149. https://doi.org/10.1016/j.isprsjprs.2017.05.002

Ampatzidis, Y., Partel, V., 2019. UAV-based high throughput phenotyping in citrus utilizing multispectral imaging and artificial intelligence. Remote Sens., 11(4), 410-429. https://doi.org/10.3390/rs11040410

An, J., Li, W., Li, M., Cui, S., Yue, H., 2019. Identification and classification of maize drought stress using deep convolutional neural network. Symmetry, 11(2), 1–14. https://doi.org/10.3390/sym11020256

Bah, M.D., Hafiane, A., Canals, R., 2020. CRowNet: Deep Network for Crop Row Detection in UAV Images. IEEE Access 8, 5189–5200. https://doi.org/10.1109/ACCESS.2019.2960873

Badrinarayanan, V., Kendall, A., Cipolla, R., 2017. SegNet: A deep convolutional encoder-decoder architecture for image segmentation. IEEE Trans. Pattern Anal. Mach. Intell., 39(12), 2481–2495. https://doi.org/10.1109/TPAMI.2016.2644615

Ball, J. E., Anderson, D. T., Chan, C. S. (2017). Comprehensive survey of deep learning in remote sensing: theories, tools, and challenges for the community. J. Appl. Remote Sens., 11(04), 042609. https://doi.org/10.1117/1.jrs.11.042609

Cao, Z., Simon, T., Wei, S. E., Sheikh, Y., 2017. Realtime multi-person 2D pose estimation using part affinity fields. Proc.  CVPR 2017, 1302–1310. https://doi.org/10.1109/CVPR.2017.143

Chen, S. W., Shivakumar, S. S., Dcunha, S., Das, J., Okon, E., Qu, C., … Kumar, V., 2017. Counting apples and oranges with deep learning: A data-driven approach. IEEE Robot. Autom. Lett., 2(2), 781–788. https://doi.org/10.1109/LRA.2017.2651944

Csillik, O., Cherbini, J., Johnson, R., Lyons, A., Kelly, M., 2018. Identification of citrus trees from unmanned aerial vehicle imagery using convolutional neural networks. Drones, 2(4), 39-55. https://doi.org/10.3390/drones2040039

Delloye, C., Weiss, M., Defourny, P., 2018. Retrieval of the canopy chlorophyll content from Sentinel-2 spectral bands to estimate nitrogen uptake in intensive winter wheat cropping systems. Remote Sens. Environ., 216, 245–261. https://doi.org/10.1016/j.rse.2018.06.037

Deng, L., Mao, Z., Li, X., Hu, Z., Duan, F., Yan, Y., 2018. UAV-based multispectral remote sensing for precision agriculture: A comparison between different cameras. ISPRS J. Photogramm. Remote Sens.,146, 124–136. https://doi.org/10.1016/j.isprsjprs.2018.09.008

Dian Bah, M., Hafiane, A., Canals, R., 2018. Deep learning with unsupervised data labeling for weed detection in line crops in UAV images. Remote Sens., 10(11), 1–20. https://doi.org/10.3390/rs10111690

Djerriri, K., Ghabi, M., Karoui, M. S., Adjoudj, R., 2018. Palm trees counting in remote sensing imagery using regression convolutional neural network. Proc. IGARSS, 2627–2630. https://doi.org/10.1109/IGARSS.2018.8519188

Fan, Z., Lu, J., Gong, M., Xie, H., Goodman, E. D., 2018. Automatic tobacco plant detection in UAV images via deep neural networks. IEEE J. Sel. Top. Appl. Earth Observ. Remote Sens., 11(3), 876–887. https://doi.org/10.1109/JSTARS.2018.2793849

Freudenberg, M., Nölke, N., Agostini, A., Urban, K., Wörgötter, F., Kleinn, C., 2019. Large scale palm tree detection in high resolution satellite images using U-Net. Remote Sens., 11(3), 1–18. https://doi.org/10.3390/rs11030312

Ghamisi, P., Plaza, J., Chen, Y., Li, J., Plaza, A. J., 2017. Advanced spectral classifiers for hyperspectral images: A review. IEEE Geosci. Remote Sens. M.,  5(1), 8–32. https://doi.org/10.1109/MGRS.2016.2616418

Gnädinger, F., Schmidhalter, U., 2017. Digital counts of maize plants by Unmanned Aerial Vehicles (UAVs). Remote Sens. 9. https://doi.org/10.3390/rs9060544

Goldman, E., Herzig, R., Eisenschtat, A., Ratzon, O., Levi, I., Goldberger, J., Hassner, T., 2019. Precise detection in densely packed scenes. Proc. CVPR, https://doi.org/10.1109/CVPR.2019.00537

Hartling, S., Sagan, V., Sidike, P., Maimaitijiang, M., Carron, J., 2019. Urban tree species classification using a WorldView-2/3 and LiDAR data fusion approach and deep learning. Sensors, 19(6), 1–23. https://doi.org/10.3390/s19061284

Hassanein, M., Khedr, M., El-Sheimy, N., 2019. Crop row detection procedure using low-cost UAV imagery system. ISPRS Archives, 42(2/W13), 349–356. https://doi.org/10.5194/isprs-archives-XLII-2-W13-349-2019

Ho Tong Minh, D., Ienco, D., Gaetano, R., Lalande, N., Ndikumana, E., Osman, F., Maurel, P., 2018. Deep recurrent neural networks for winter vegetation quality mapping via multitemporal SAR Sentinel-1. IEEE Geosci. Remote Sens. Lett., 15(3), 465–468. https://doi.org/10.1109/LGRS.2018.2794581

Huang, X.; Liu, X.; Zhang, L. 2014. A Multichannel Gray Level Co-Occurrence Matrix for Multi/Hyperspectral Image Texture Representation. Remote Sens., 6, 8424-8445. https://doi.org/10.3390/rs6098424

Hunt, E. R., Daughtry, C. S. T., 2018. What good are unmanned aircraft systems for agricultural remote sensing and precision agriculture? Int. J. Remote Sens., 39(15–16), 5345–5376. https://doi.org/10.1080/01431161.2017.1410300

Hunt, M. L., Blackburn, G. A., Carrasco, L., Redhead, J. W., and Rowland, C. S., 2019. High resolution wheat yield mapping using Sentinel-2. Remote Sens. Environ.,  233, 111410. https://doi.org/10.1016/j.rse.2019.111410

Hussain, M., Chen, D., Cheng, A., Wei, H., and Stanley, D. 2013. Change detection from remotely sensed images: From pixel-based to object-based approaches. ISPRS Journal of Photogrammetry and Remote Sensing, 80, 91–106. https://doi.org/10.1016/j.isprsjprs.2013.03.006

Ioffe, S., and Szegedy, C., 2015. Batch normalization: Accelerating deep network training by reducing internal covariate shift. Proc. 32nd Int. Conf. Machine Learning (ICML) 1, 448–456.

Jakubowski, M. K., Li, W., Guo, Q., and Kelly, M., 2013. Delineating individual trees from lidar data: A comparison of vector- and raster-based segmentation approaches. Remote Sens., 5(9), 4163–4186. https://doi.org/10.3390/rs5094163

Jensen, J. R. 2015. Introductory Digital Image Processing: A Remote Sensing Perspective. 4th. Pearson Series in Geographic Information Science. 659p.

Jiang, H., Chen, S., Li, D., Wang, C., and Yang, J., 2017. Papaya Tree detection with UAV images using a GPU-accelerated scale-space filtering method. Remote Sens., 9(7), 721-734. https://doi.org/10.3390/rs9070721

Jiang, Y., Li, C., Paterson, A.H., Robertson, J.S., 2019. DeepSeedling: Deep convolutional network and Kalman filter for plant seedling detection and counting in the field. Plant Methods 15, 1–19. https://doi.org/10.1186/s13007-019-0528-3

Jin, S., Su, Y., Gao, S., Wu, F., Hu, T., Liu, J., Li, W., Wang, D., Chen, S., Jiang, Y., Pang, S., Guo, Q., 2018. Deep learning: Individual maize segmentation from terrestrial lidar data using faster R-CNN and regional growth algorithms. Front. Plant Sci. 9, 1–10. https://doi.org/10.3389/fpls.2018.00866

Jin, Z., Azzari, G., You, C., Di Tommaso, S., Aston, S., Burke, M., and Lobell, D. B., 2019. Smallholder maize area and yield mapping at national scales with Google Earth Engine. Remote Sens. Environ.,  228, 115–128. https://doi.org/10.1016/j.rse.2019.04.016

Kamilaris, A., and Prenafeta-Boldú, F. X., 2018. Deep learning in agriculture: A survey. Comput. Electron. Agric., 147, 70–90. https://doi.org/10.1016/j.compag.2018.02.016

Khamparia, A., and Singh, K. M., 2019. A systematic review on deep learning architectures and applications. Expert Syst., 36(3), 1–22. https://doi.org/10.1111/exsy.12400

Kitano, B. T., Mendes, C. C. T., Geus, A. R., Oliveira, H. C., and Souza, J. R., 2019. Corn plant counting using deep learning and UAV images. IEEE Geosci. Remote Sens. Lett., 1–5. https://doi.org/10.1109/lgrs.2019.2930549

Krizhevsky, A., 2014. One weird trick for parallelizing convolutional neural networks. arXiv:1404.5997 http://arxiv.org/abs/1404.5997

Larsen, M., Eriksson, M., Descombes, X., Perrin, G., Brandtberg, T., and Gougeon, F. A., 2011. Comparison of six individual tree crown detection algorithms evaluated under varying forest conditions. Int. J. Remote Sens., 32(20), 5827–5852. https://doi.org/10.1080/01431161.2010.507790

Lecun, Y., Bengio, Y., and Hinton, G., 2015. Deep learning. Nature, 521(7553), 436–444. https://doi.org/10.1038/nature14539

Leiva, J. N., Robbins, J., Saraswat, D., She, Y., and Ehsani, R., 2017. Evaluating remotely sensed plant count accuracy with differing unmanned aircraft system altitudes, physical canopy separations, and ground covers. J. Appl. Remote Sens.,11(3), 036003. https://doi.org/10.1117/1.jrs.11.036003

Li, D., Guo, H., Wang, C., Li, W., Chen, H., and Zuo, Z., 2016. Individual tree delineation in windbreaks using airborne-laser-scanning data and unmanned aerial vehicle stereo images. IEEE Geosci. Remote Sens. Lett., 13(9), 1330–1334. https://doi.org/10.1109/LGRS.2016.2584109

Li, W., Fu, H., Yu, L., and Cracknell, A., 2017. Deep learning based oil palm tree detection and counting for high-resolution remote sensing images. Remote Sens., 9(1), 22-35. https://doi.org/10.3390/rs9010022

Lin, T. Y., Goyal, P., Girshick, R., He, K., and Dollar, P., 2017. Focal loss for dense object detection. Proc. CVPR, 2999–3007. https://doi.org/10.1109/ICCV.2017.324

Liu, Y., Gross, L., Li, Z., Li, X., Fan, X., and Qi, W., 2019. Automatic building extraction on high-resolution remote sensing imagery using deep convolutional encoder-decoder with spatial pyramid pooling. IEEE Access, 7, 128774–128786. https://doi.org/10.1109/access.2019.2940527

Miyoshi, G.T., Arruda, M. dos S., Osco, L.P., Junior, J.M., Gonçalves, D.N., Imai, N.N., Tommaselli, A.M.G., Honkavaara, E., Gonçalves, W.N., 2020. A novel deep learning method to identify single tree species in UAV-based hyperspectral images. Remote Sens. 12. https://doi.org/10.3390/RS12081294

Mochida, K., Koda, S., Inoue, K., Hirayama, T., Tanaka, S., Nishii, R., and Melgani, F. 2018. Computer vision-based phenotyping for improvement of plant productivity: A machine learning perspective. GigaScience, 8(1), 1–12. https://doi.org/10.1093/gigascience/giy153

Mohanty, S. K., and Swain, M. R., 2019. Bioethanol production from corn and wheat: Food, fuel, and future. in bioethanol production from food crops. Bioethanol Production from Food Crops Sustainable Sources, Interventions, and Challenges, https://doi.org/10.1016/b978-0-12-813766-6.00003-5

Ndikumana, E., Minh, D. H. T., Baghdadi, N., Courault, D., and Hossard, L., 2018. Deep recurrent neural network for agricultural classification using multitemporal SAR Sentinel-1 for Camargue, France. Remote Sens., 10(8), 1–16. https://doi.org/10.3390/rs10081217

Nevalainen, O., Honkavaara, E., Tuominen, S., Viljanen, N., Hakala, T., Yu, X., … Tommaselli, A. M. G., 2017. Individual tree detection and classification with UAV-Based photogrammetric point clouds and hyperspectral imaging. Remote Sens., 9(3), 185-219. https://doi.org/10.3390/rs9030185

Oliveira, H. C., Guizilini, V. C., Nunes, I. P., and Souza, J. R., 2018. Failure detection in row crops from UAV images using morphological operators. IEEE Geosci. Remote Sens. Lett.,15(7), 991–995. https://doi.org/10.1109/LGRS.2018.2819944

Osco, L. P., Paula, A., Ramos, M., Pereira, D. R., Akemi, É., Moriya, S., … Matsubara, E. T., 2019. Predicting canopy nitrogen content in citrus-trees using random forest algorithm associated to spectral vegetation indices from UAV-imagery. Remote Sens., 2019, 11(24), 2925-2942; https://doi.org/10.3390/rs11242925

Osco, L. P., Ramos, A. P. M., Pinheiro, M. M. F., Moriya, É. A. S., Imai, N. N., Estrabis, N., Ianczyk, F., de Araújo, F. F., Liesenberg, V., de Castro Jorge, L. A., Li, J., Ma, L., Gonçalves, W. N., Marcato, J., and Creste, J. E. (2020a). A machine learning framework to predict nutrient content in valencia-orange leaf hyperspectral measurements. Remote Sensing, 12(6). https://doi.org/10.3390/rs12060906

Osco, L. P., Arruda, M. S., Junior, J. M., da Silva, N. B., Ramos, A. P. M., Moriya, É. A. S., ... Li, J., 2020b. A convolutional neural network approach for counting and geolocating citrus-trees in UAV multispectral imagery. ISPRS J. Photogramm. Remote Sens., 160, 97-106. https://doi.org/10.1016/j.isprsjprs.2019.12.010

Özcan, A. H., Hisar, D., Sayar, Y., and Ünsalan, C., 2017. Tree crown detection and delineation in satellite images using probabilistic voting. Remote Sens. Lett., 8(8), 761–770. https://doi.org/10.1080/2150704X.2017.1322733

Ozdarici-Ok, A., 2015. Automatic detection and delineation of citrus trees from VHR satellite imagery. Int. J. Remote Sens., 36(17), 4275–4296. https://doi.org/10.1080/01431161.2015.1079663

Paoletti, M. E., Haut, J. M., Plaza, J., and Plaza, A., 2018. A new deep convolutional neural network for fast hyperspectral image classification. ISPRS J. Photogramm. Remote Sens., 145, 120–147. https://doi.org/10.1016/j.isprsjprs.2017.11.021

Primicerio, J., Caruso, G., Comba, L., Crisci, A., Gay, P., Guidoni, S., … Vaccari, F. P., 2017. Individual plant definition and missing plant characterization in vineyards from high-resolution UAV imagery. European J. Remote Sens., 50(1), 179–186. https://doi.org/10.1080/22797254.2017.1308234

Quan, L., Feng, H., Lv, Y., Wang, Q., Zhang, C., Liu, J., Yuan, Z., 2019. Maize seedling detection under different growth stages and complex field environments based on an improved Faster R–CNN. Biosyst. Eng. 184, 1–23. https://doi.org/10.1016/j.biosystemseng.2019.05.002

Redmon, J., and Farhadi, A., 2018. YOLOv3: An incremental improvement. Retrieved from http://arxiv.org/abs/1804.02767

Ren, S., He, K., Girshick, R., Sun, J., 2015. Faster R-CNN: Towards real-time object detection with region proposal networks. Proc. NIPS , 28, 91–99.

Ribera, J., Chen, Y., Boomsma, C., and Delp, E. J., 2018. Counting plants using deep learning. Prof. GlobalSIP 2017, 1344–1348. https://doi.org/10.1109/GlobalSIP.2017.8309180

Ronneberger, O., Fischer, P., and Brox, T., 2015. U-net: Convolutional networks for biomedical image segmentation. Lecture Notes in Computer Science (Including Subseries Lecture Notes in Artificial Intelligence and Lecture Notes in Bioinformatics), 9351, 234–241. https://doi.org/10.1007/978-3-319-24574-4-28

Safonova, A., Tabik, S., Alcaraz-Segura, D., Rubtsov, A., Maglinets, Y., and Herrera, F., 2019. Detection of fir trees (Abies sibirica) damaged by the bark beetle in unmanned aerial vehicle images with deep learning. Remote Sens., 11(6), 643-462. https://doi.org/10.3390/rs11060643

Salamí, E., Gallardo, A., Skorobogatov, G., and Barrado, C.,2019. On-the-fly olive tree counting using a UAS and cloud services. Remote Sens.,  11(3), 316-337. https://doi.org/10.3390/rs11030316

Santos, A. A., Marcato Junior, J., Araújo, M. S., Di Martini, D. R., Tetila, E. C., Siqueira, H. L., … Gonçalves, W. N, 2019. Assessment of CNN-based methods for individual tree detection on images captured by RGB cameras attached to UAVs. Sensors, 19(16), 1–11. https://doi.org/10.3390/s19163595

Simonyan, K., and Zisserman, A., 2015. Very deep convolutional networks for large-scale image recognition. Proc. ICLR 2015, arXiv:1409.1556.

Sun, J., Di, L., Sun, Z., Shen, Y., and Lai, Z. (2019). County-level soybean yield prediction using deep CNN-LSTM model. Sensors, 19(20), 1–21. https://doi.org/10.3390/s19204363

Sylvain, J. D., Drolet, G., and Brown, N. (2019). Mapping dead forest cover using a deep convolutional neural network and digital aerial photography. ISPRS J. Photogramm. Remote Sens., 156, 14–26. https://doi.org/10.1016/j.isprsjprs.2019.07.010

Szegedy, C., Vanhoucke, V., Ioffe, S., Shlens, J., and Wojna, Z., 2016. Rethinking the inception architecture for computer vision. Proc. CVPR, 2818–2826. https://doi.org/10.1109/CVPR.2016.308

Szegedy, C., Ioffe, S., Vanhoucke, V., and Alemi, A. A., 2017. Inception-v4, inception-ResNet and the impact of residual connections on learning. Proc. AAAI 2017, 4278–4284.

Tao, S., Wu, F., Guo, Q., Wang, Y., Li, W., Xue, B., … Fang, J., 2015. Segmenting tree crowns from terrestrial and mobile LiDAR data by exploring ecological theories. ISPRS J. Photogramm. Remote Sens., 110, 66–76. https://doi.org/10.1016/j.isprsjprs.2015.10.007

Varela, S., Dhodda, P. R., Hsu, W. H., Prasad, P. V. V., Assefa, Y., Peralta, N. R., … Ciampitti, I. A., 2018. Early-season stand count determination in corn via integration of imagery from unmanned aerial systems (UAS) and supervised learning techniques. Remote Sens., 10(2), 343-357. https://doi.org/10.3390/rs10020343

Verma, N. K., Lamb, D. W., Reid, N., and Wilson, B., 2016. Comparison of canopy volume measurements of scattered eucalypt farm trees derived from high spatial resolution imagery and LiDAR. Remote Sens., 8(5), 388-404. https://doi.org/10.3390/rs8050388

Wang, H., Magagi, R., Goïta, K., Trudel, M., McNairn, H., and Powers, J., 2019a. Crop phenology retrieval via polarimetric SAR decomposition and Random Forest algorithm. Remote Sens. Environ.,  231, 111234. https://doi.org/10.1016/j.rse.2019.111234

Wang, S., Azzari, G., and Lobell, D. B., 2019b. Crop type mapping without field-level labels: Random forest transfer and unsupervised clustering techniques. Remote Sens. Environ., 222, 303–317. https://doi.org/10.1016/j.rse.2018.12.026

Weinstein, B. G., Marconi, S., Bohlman, S., Zare, A., and White, E., 2019. Individual tree-crown detection in RGB imagery using semi-supervised deep learning neural networks. Remote Sens., 11(11), 1309-1322. https://doi.org/10.3390/rs11111309

Weiss, M., Jacob, F., and Duveiller, G. 2020. Remote sensing for agricultural applications: A meta-review. Remote Sens. Environ., 236(November 2019), 111402. https://doi.org/10.1016/j.rse.2019.111402

Wu, J., Yang, G., Yang, X., Xu, B., Han, L., and Zhu, Y., 2019. Automatic counting of in situ rice seedlings from UAV images based on a deep fully convolutional neural network. Remote Sens., 11(6), 691-710. https://doi.org/10.3390/rs11060691

Zhang, T. Y., and Suen, C. Y., 1984. A fast parallel algorithm for thinning digital patterns. Communications of the ACM, 27(3), 236-239.

Zhang, H., Li, Y., Zhang, Y., and Shen, Q., 2017. Spectral-spatial classification of hyperspectral imagery using a dual-channel convolutional neural network. Remote Sens. Lett., 8(5), 438–447. https://doi.org/10.1080/2150704X.2017.1280200

Zhao, H., Shi, J., Qi, X., Wang, X., and Jia, J., 2017. Pyramid scene parsing network. Proc. CVPR, 2881-2890.

Zhong, L., Hu, L., and Zhou, H., 2019. Deep learning based multi-temporal crop classification. Remote Sens. Environ., 221, 430–443. https://doi.org/10.1016/j.rse.2018.11.032

Zhou, C., Yang, G., Liang, D., Yang, X., Xu, B., 2018. An integrated skeleton extraction and pruning method for spatial recognition of maize seedlings in MGV and UAV remote images. IEEE Trans. Geosci. Remote Sens. 56, 4618–4632. https://doi.org/10.1109/TGRS.2018.2830823



\end{document}